\documentclass[letterpaper]{article} 
\usepackage{aaai2026}  
\usepackage{times}  
\usepackage{helvet}  
\usepackage{courier}  
\usepackage[hyphens]{url}  
\usepackage{graphicx} 
\usepackage{natbib}  
\usepackage{caption} 
\usepackage{algorithm}
\usepackage{algorithmic}

\usepackage{multirow}
\usepackage{booktabs}
\usepackage{subcaption}
\usepackage[table]{xcolor}
\usepackage{amsfonts}       
\usepackage{nicefrac}       
\usepackage{microtype}      
\usepackage{xcolor}  
\usepackage{colortbl}
\usepackage{amsmath} 
\usepackage{caption}
\usepackage{pifont}
\usepackage{arydshln}
\usepackage{fontawesome}
\usepackage{svg}
\usepackage{graphicx}
\usepackage{tcolorbox}  
\usepackage{adjustbox}  
\usepackage{enumitem}
\usepackage[flushleft]{threeparttable}
\usepackage{newfloat}
\usepackage{listings}
\DeclareCaptionStyle{ruled}{labelfont=normalfont,labelsep=colon,strut=off} 
\floatstyle{ruled}
\newfloat{listing}{tb}{lst}{}
\floatname{listing}{Listing}
\tcbset{
  promptstyle/.style={
    colback=blue!5,  
    colframe=blue,
    boxrule=1pt,     
    arc=4pt,          
    left=6pt,right=6pt,top=6pt,bottom=6pt,
    boxsep=4pt,    
    before upper={\parindent15pt},
    overlay={
      \fill[blue!30!white] (frame.south west) -- ++(0.5cm,0) -- ++(0,-0.5cm) -- cycle;
      \fill[blue!30!white] (frame.north east) -- ++(-0.5cm,0) -- ++(0,0.5cm) -- cycle;
    }
  }
}

\newtcolorbox{promptblock}[1][]{
  promptstyle,
  #1
}
\usepackage{graphicx}
\usepackage{xspace}

\newcommand{\github}{\raisebox{-1.5pt}{\includegraphics[height=1.05em]{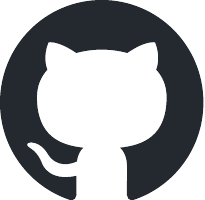}}\xspace}
\newcommand{\worldwideweb}{\raisebox{-1.5pt}{\includegraphics[height=1.05em]{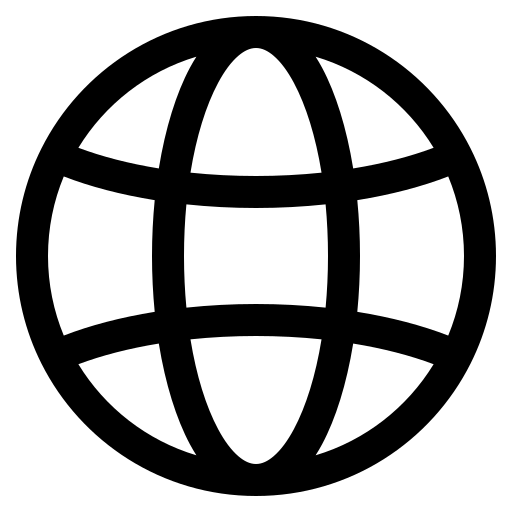}}\xspace}

\def\mdname{UniME-V2}
\def\mdrkname{UniME-V2-Reranker}
\definecolor{kcgreen}{rgb}{0.1, 0.7, 0.2}
\definecolor{kcred}{rgb}{139, 0, 0}
\definecolor{green_ours}{HTML}{D8ECD1}
\definecolor{color1}{RGB}{157,27,229}
\definecolor{color2}{RGB}{127,220,244}
\usepackage{xcolor}  
\definecolor{mygreen}{rgb}{0.1, 0.7, 0.2}
\definecolor{myred}{rgb}{139, 0, 0}
\definecolor{myorange}{rgb}{0.93, 0.51, 0.18}
\definecolor{LightCyan}{rgb}{0.96,0.96,0.96}
\definecolor{lightblue}{RGB}{240,240,240}

\title{UniME-V2: MLLM-as-a-Judge for Universal Multimodal \\ Embedding Learning}
\author{Tiancheng Gu$^{\text{\ding{170}, \ding{171}} \equalcontrib}$, Kaicheng Yang$^{\clubsuit \equalcontrib}$, 
Kaichen Zhang$^{\text{\ding{170}, \ding{169}}}$, 
Xiang An$^{\clubsuit}$, 
Ziyong Feng$^{\clubsuit}$, \\ 
Yueyi Zhang$^{\text{\ding{170}}}$\textsuperscript{$\ddagger$},
Weidong Cai$^{\text{\ding{171}}}$, 
Jiankang Deng$^{\text{\ding{90}}}$\textsuperscript{$\ddagger$}, 
Lidong Bing$^{\text{\ding{170}}}$
}
\affiliations{
    $^{\text{\ding{170}}}$MiroMind AI
    $^{\text{\ding{171}}}$The University of Sydney 
    $^{\clubsuit}$M.R.L. Team \\
    $^{\text{\ding{169}}}$LMMs-Lab Team
    $^{\text{\ding{90}}}$Imperial College London \\
    \vspace{1mm}
    \texttt\tiny{yueyi.zhang@miromind.ai, j.deng16@imperial.ac.uk} \\
    \vspace{1.5mm}
    \worldwideweb \ Webpage:  \underline{https://garygutc.github.io/UniME-v2} \\
    \github \ Github: \underline{https://github.com/GaryGuTC/UniME-v2}
}

\usepackage{bibentry}

\begin{document}

\maketitle

\begin{abstract}
Universal multimodal embedding models are foundational to various tasks. Existing approaches typically employ in-batch negative mining by measuring the similarity of query-candidate pairs. However, these methods often struggle to capture subtle semantic differences among candidates and lack diversity in negative samples. Moreover, the embeddings exhibit limited discriminative ability in distinguishing false and hard negatives. In this paper, we leverage the advanced understanding capabilities of MLLMs to enhance representation learning and present a novel \textbf{Uni}versal \textbf{M}ultimodal \textbf{E}mbedding~(\textbf{UniME-V2}) model. Our approach first constructs a potential hard negative set through global retrieval. We then introduce the MLLM-as-a-Judge mechanism, which utilizes MLLMs to assess the semantic alignment of query-candidate pairs and generate soft semantic matching scores. These scores serve as a foundation for hard negative mining, mitigating the impact of false negatives and enabling the identification of diverse, high-quality hard negatives. Furthermore, the semantic matching scores are used as soft labels to mitigate the rigid one-to-one mapping constraint. By aligning the similarity matrix with the soft semantic matching score matrix, the model learns semantic distinctions among candidates, significantly enhancing its discriminative capacity. To further improve performance, we propose \textbf{\mdrkname}, a reranking model trained on our mined hard negatives through a joint pairwise and listwise optimization approach. We conduct comprehensive experiments on the MMEB benchmark and multiple retrieval tasks, demonstrating that our method achieves state-of-the-art performance on average across all tasks. 

\end{abstract}
\section{Introduction}

\begin{figure}[t!]
\centering
\includegraphics[width=0.98\linewidth]{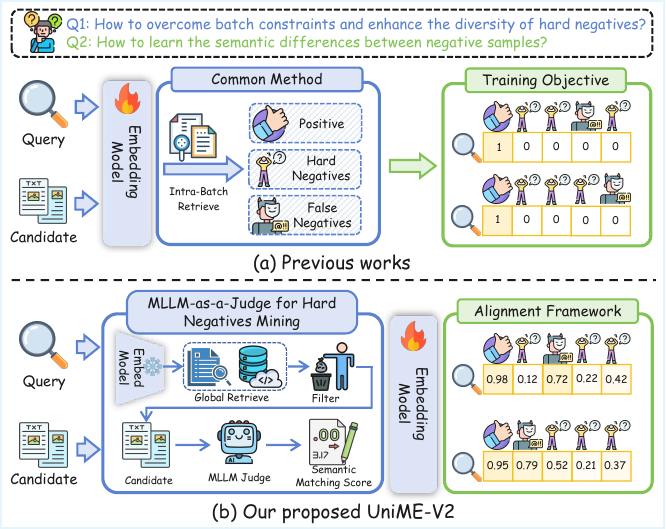}
\vspace{-2mm}
\caption{Comparison between previous works and \mdname. \mdname\ exploits the understanding capabilities of MLLMs for hard negatives mining and generates a soft semantic matching score to supervise the model in learning the semantic difference among candidates.}
\vspace{-4mm}
\label{fig:different}
\end{figure}

Multimodal embedding models aim to encode heterogeneous multimodal data into a unified dense representation space, enabling a wide range of downstream applications such as visual question answering~\cite{dong2024toward, hamza2025llava,li2025miv} and multimodal retrieval~\cite{zheng2025gradient,gu2025realsyn,yang2025clip,gu2024rwkv}. With the increasing adoption of these models, multimodal representation learning has garnered significant research attention. Among these models, CLIP~\cite{CLIP} stands out as a pioneering approach, achieving remarkable performance in text-image retrieval by leveraging cross-modal contrastive learning on large-scale web-collected image-text pairs~\cite{schuhmann2021laion}. However, its effectiveness is hindered by three major limitations: (1) CLIP enforces a strict text token limit of 77, which restricts its ability to process detailed or lengthy descriptions~\cite{longclip,flame,llm2clip}; (2) Its dual-encoder design processes images and text independently, which reduces its effectiveness in handling complex tasks, such as instruction-following multimodal retrieval~\cite{VLM2Vec, LamRA, unime}; and (3) CLIP exhibits limited proficiency in advanced language understanding, struggles with handling compositionality, and often demonstrates bag-of-words behavior~\cite{negclip,tschannen2023image,degla}.

Recent advances in Large Language Models (LLMs) have achieved state-of-the-art performance on the MTEB benchmark~\cite{mteb}. Motivated by these developments~\cite{nvembed,llm2vec}, researchers are currently exploring how to utilize Multimodal Large Language Models (MLLMs) to learn universal multimodal representation. E5-V~\cite{E5V} adopts a unimodal contrastive learning approach, training the language component of MLLMs on sentence pairs to better align the cross-modal representation spaces. VLM2Vec~\cite{VLM2Vec} introduces the Massive Multimodal Embedding Benchmark (MMEB), comprising 36 datasets across four meta-tasks, and proposes a contrastive learning framework to repurpose pre-trained vision-language models as embedding models via training on the MMEB dataset. QQMM~\cite{xue2025improve} provides an in-depth analysis of the gradients derived from the InfoNCE loss and proposes amplifying gradients associated with hard negative samples to encourage the model to learn more discriminative embeddings. UniME~\cite{unime} presents a two-stage framework that leverages a powerful LLM-based teacher model to improve the embedding capabilities of the language component in MLLMs. Furthermore, it incorporates a hard negative sampling strategy that selects multiple challenging negatives per instance within each batch. Despite these advances, existing methods fail to fully exploit the semantic differences between candidates and are limited by the lack of diversity in negative samples. Moreover, the raw embeddings produced by these models are frequently inadequate for reliably discriminating between hard negatives and false negatives.

In this paper, we propose a novel \textbf{Uni}versal \textbf{M}ultimodal \textbf{E}mbedding~(\textbf{\mdname}) model, which leverages the robust understanding capabilities of MLLMs to enhance representation learning. As shown in Fig.~\ref{fig:different}, we first construct a potential hard negative set through global retrieval. Then, we introduce MLLM-as-a-Judge to assess the semantic alignment of query-candidate pairs, producing semantic matching scores. This score serves as the foundation for hard negative mining, effectively reducing interference from false negatives and enabling the identification of high-quality, diverse hard negatives. Additionally, we use the scores as soft labels to mitigate strict one-to-one mapping constraints. Aligning the similarity matrix with the semantic score matrix enables the model to capture semantic distinctions among candidates, significantly improving its discriminative ability. To further enhance performance, we introduce \textbf{\mdrkname}, a reranking model trained on our mined hard negatives through a joint pairwise and listwise optimization approach. Extensive experiments on the MMEB benchmark and various retrieval tasks, including short/long caption retrieval and compositional retrieval, demonstrate that our method achieves state-of-the-art performance across all tasks. The main contributions of this paper are summarized as follows:
\begin{itemize}[leftmargin=*]
    \item We introduce \textbf{a MLLM-as-a-Judge pipeline for hard negative mining} that uses the advanced understanding capabilities of MLLM to assess the semantic alignment of each query-candidate pair within a globally retrieved potential hard negative set.
    \item We present \textbf{\mdname}, a novel universal multimodal embedding model trained with \textbf{an MLLM judgment based distribution alignment framework}. By leveraging semantic matching scores as soft labels, the model effectively captures semantic differences between candidates, significantly enhancing its discriminative capability.
    \item We propose \textbf{\mdrkname}, a reranking model trained on high-quality, diverse hard negatives through a joint pairwise and listwise optimization approach.
    \item We conduct \textbf{extensive experiments} on the MMEB benchmark and various retrieval tasks, including short and long caption retrieval as well as compositional retrieval. The results demonstrate that our method achieves state-of-the-art performance on average across all tasks.
 \end{itemize}
\section{Related Work}
\subsection{Multimodal Large Language Models}
Multimodal Large Language Models (MLLMs) extend traditional LLMs to process and integrate information across multiple modalities~\cite{wang2025vl,wang2025code,xie2024croc,bai2025qwen2,an2025unictokens, an2024mc}. As a foundational contribution, LLaVA~\cite{llava} leverages a subset of the CC3M~\cite{cc3m} dataset to achieve more balanced conceptual coverage. In this approach, the visual encoder and language model remain frozen, while only the projection layer is trained to align visual features with language tokens. Subsequently, numerous MLLM variants~\cite{kosmos,vila,tang2025intervening,tang2025seeing,an2025llava} have achieved remarkable results in multimodal understanding and reasoning tasks. For instance, CogVLM~\cite{wang2023cogvlm} incorporates a trainable visual expert module into the attention and feed-forward layers of the language model, achieving significant improvements on 17 standard cross-modal benchmarks. Similarly, Qwen2-VL~\cite{wang2024qwen2} introduces the Naive Dynamic Resolution mechanism and integrates M-RoPE to enhance positional information fusion, yielding competitive performance across diverse benchmarks. LLaVA-OneVision~\cite{li2024llava} pushes the boundaries of open MLLMs by excelling in single-image, multi-image, and video tasks, showcasing robust video understanding through effective task transfer from image-based training. Although these advances have significantly improved the understanding capabilities of MLLMs, further research is needed to explore how MLLMs can effectively learn unified multimodal representations.

\begin{figure*}[t!]
    \centering
    \includegraphics[width=0.95\linewidth]{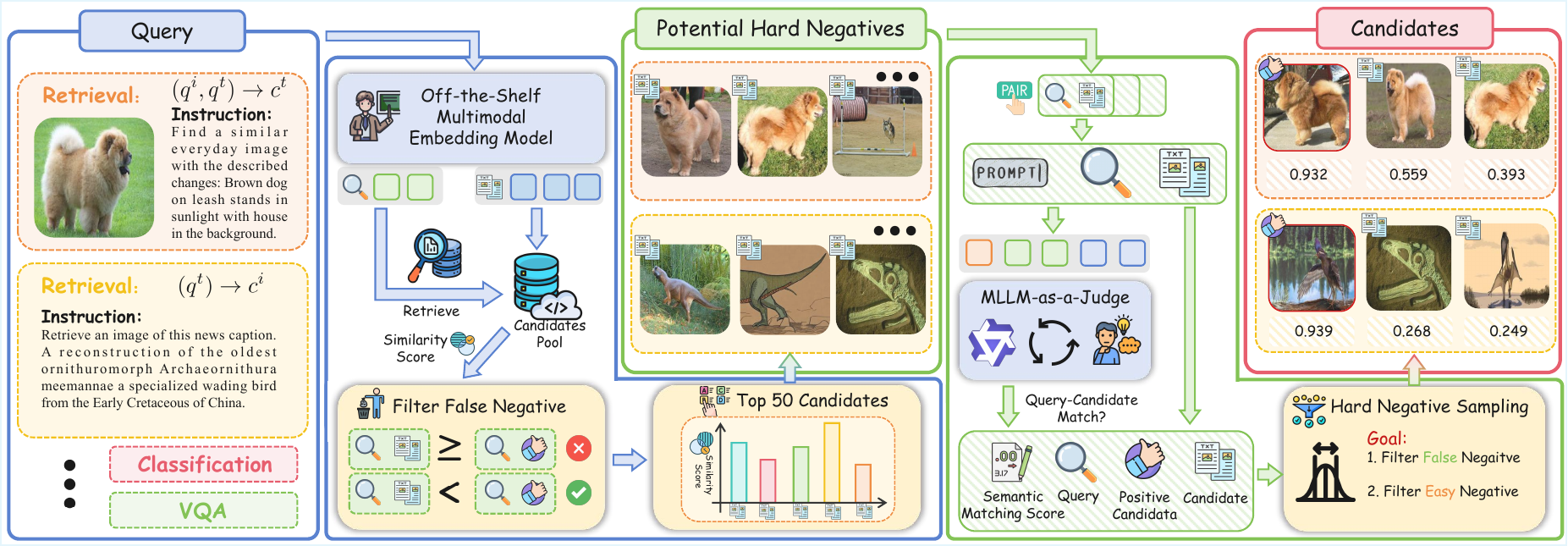}
    \vspace{-1mm}
    \caption{The MLLM-as-a-Judge pipeline for Hard Negatives Mining. We first utilize an existing multimodal embedding model for global retrieval to construct a potential hard negative set. We then leverage the powerful understanding capabilities of MLLM to score query-candidate pairs based on their semantic alignment, enabling precise identification of hard negatives.}
    \vspace{-3mm}
    \label{fig:data_pipeline}
\end{figure*}

\subsection{Multimodal Representation Learning}
CLIP~\cite{CLIP} demonstrates strong image-text retrieval performance through large-scale cross-modal contrastive learning but faces three key limitations: (1) A 77-token text truncation restricts fine-grained semantic alignment~\cite{longclip,flame,llm2clip}; (2) Its dual-encoder architecture limits effective cross-modal fusion, particularly for instruction-sensitive tasks~\cite{VLM2Vec, LamRA, unime}; and (3) Simplistic language modeling results in bag-of-words representations~\cite{negclip,tschannen2023image,degla,lei2023mcad, huang2024cross, andonian2022robust}. To address these issues, recent studies have incorporated MLLMs for enhanced multimodal representation learning. E5-V~\cite{E5V} employs unimodal contrastive learning, training the language component of MLLMs on sentence pairs to reduce cross-modal representation gaps. VLM2Vec~\cite{VLM2Vec} introduces the Massive Multimodal Embedding Benchmark (MMEB) and adapts state-of-the-art vision-language models into embedding models using a contrastive framework trained on MMEB. QQMM~\cite{xue2025improve} analyzes InfoNCE loss gradients and proposes enhancing gradients associated with hard negatives to improve embedding discrimination. UniME~\cite{unime} adopts a two-stage framework with an LLM-based teacher model to refine language embeddings and employs a hard negative sampling strategy, selecting multiple challenging negatives per batch. Despite these advancements, existing methods still under-utilize the semantic differences among candidates and struggle to effectively identify and leverage hard negatives during retrieval.
\section{Methodology}

\subsection{Task Definition}
Unlike CLIP, which employs separate encoders to generate embeddings for each modality, we investigate leveraging the unified architecture of MLLM to extract embeddings across multiple modalities and improve retrieval performance through reranking. Specifically, given a query $q$ and a set of candidates $\Omega_c = \{c_1, c_2, \dots, c_n\}$, which may include images, text, and interleaved image-text data, the universal embedding model $ \Phi_{\text{emb}}$ encodes the query and candidates, retrieving the top-$k$ most relevant candidates $\Omega_k = \Phi_{\text{emb}}(q, \Omega_c)$. To further enhance retrieval performance, a reranker model $\Phi_{\text{rank}}$ refines this subset through a reranking process, producing the final ranked output $\hat{\Omega}_k = \Phi_{\text{rank}}(q, \Omega_k)$.

\def\datapipelinename{MLLM-as-a-Judge for Hard Negatives Mining}
\subsection{\datapipelinename}
Previous works~\cite{VLM2Vec, LamRA} primarily rely on in-batch hard negative mining, where query-candidate embedding similarities are computed to sample negatives. However, this method often suffers from limited negative sample diversity and insufficient embedding discriminative power to effectively distinguish false and hard negatives. To overcome these challenges, as shown in Fig.~\ref{fig:data_pipeline}, we first construct a potential hard negative set using global retrieval. After that, inspired by previous work~\cite{zheng2023judging,chen2024mllm}, we leverage the robust understanding capabilities of MLLMs to assess the semantic alignment of each query-candidate pair and generate a soft semantic matching score. This score guides hard negative mining, enabling the identification of diverse and high-quality hard negatives while reducing the impact of false negatives.

\noindent{\textbf{Potential Hard Negative Set.}} To extract higher-quality hard negatives from global samples, we first use VLM2Vec to generate embeddings for both queries and candidates. We then retrieve the top 50 most relevant candidates for each query. To address false negatives and improve diversity, we apply a similarity threshold ($\delta$) based on the query-candidate similarity scores and select the top 50 candidates as the potential hard negative set ($\Omega_{p}$):
\begin{equation}
    \Omega_{p} = \text{Rank}_{50}\left(\left\{x_1, \dots, x_n\right\}\right), \text{where } x_i < \delta,
\end{equation}
where $x_i$ is the similarity score of the query $q$ and candidates $\hat{\Omega}_{c}$ calculated by the VLM2Vec model.

\noindent{\textbf{Semantic Matching Score.}} After constructing the potential hard negative set ($\Omega_{p})$, we employ an MLLM as a judge to compute a semantic matching score for each query-candidate pair in $\Omega_{p}$, guided by the following instruction:

\begin{promptblock}
\vspace{-2mm}
\noindent\textit{I will provide you with a query and a candidate. Please evaluate whether the candidate meets the requirements of the query. If it does, respond with 'Yes'; if it doesn't, respond with 'No'. Query:$<$Query$>$, Candidates:$<$Candidate$>$.}
\vspace{-2mm}
\end{promptblock}

After that, we compute the semantic matching score $\mathrm{S} = \{s_1, s_2, \dots, s_m\}$ based on the logits of the \texttt{Yes}~($e_y$) and \texttt{No}~($e_n$) tokens, where $s_i = \frac{e_y^i}{e_y^i + e_n^i}$, where $\mathrm{S} \in \mathbb{R}^{n_q \times 50}$ and $n_q$ denotes the number of queries. Leveraging the advanced understanding capabilities of MLLMs, the semantic matching score $\mathrm{S}$ effectively captures the degree of semantic alignment between queries and candidates.

\begin{figure}[t!]
    \centering
    \includegraphics[width=1\linewidth]{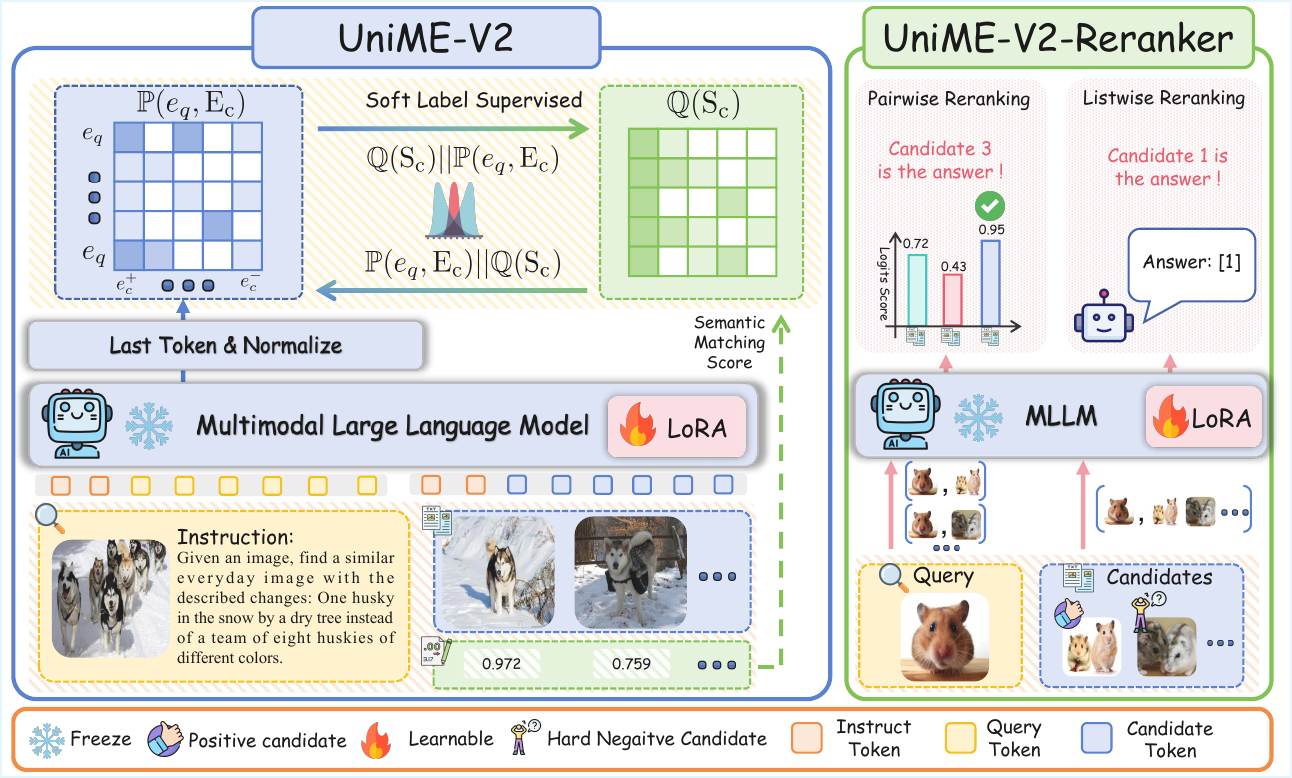}
    \vspace{-2mm}
    \caption{The architecture of the MLLM Judgment Based Training Framework. UniME-V2 uses soft semantic matching scores as supervised signals to enhance semantic distinction learning between candidates. UniME-V2-Reranker employs joint pairwise and listwise optimization to enhance reranking performance.}
    \vspace{-3mm}
    \label{fig:method2}
\end{figure}

\noindent{\textbf{Hard Negative Sampling.}} To enhance the quality of hard negatives, we refine candidates using the semantic matching score ($\mathrm{S}$). False negatives are excluded if their score exceeds a threshold $\alpha = \sigma_{q, c_{t}} - \beta$, where $c_{t}$ denotes the positive sample and $\beta$ is a hyperparameter controlling the threshold margin as 0.01. To maintain diversity, we apply a cyclical sampling strategy with five-step intervals. If the refined set contains fewer than ten candidates, we duplicate selections to ensure a minimum of ten. In the rare case where no candidates meet the criteria ($<1\%$), we randomly select 10 candidates from the initial pool of fifty and assign each a semantic matching score of 1.0. Finally, for each query $q$, we obtain the hard negative set $\Omega_h=\{c_1,...,c_k\}$ along with the corresponding semantic matching scores $\mathrm{S_h}=\{s_{q,c_1},...,s_{q,c_k}\}$.

\begin{table*}[t!]
\centering
\setlength\tabcolsep{2pt}
\renewcommand\arraystretch{1.1}
\fontsize{8pt}{8pt}\selectfont
\begin{tabular}{@{}l c cccc ccc@{}}
\toprule
\multirow{2}{*}{\textbf{Models}} 
& \multirow{2}{*}{\textbf{\#Parameters}} 
& \multicolumn{4}{c}{\textbf{Per Meta-Task Score}} 
& \multicolumn{3}{c}{\textbf{Average Score}} \\
\cmidrule(lr){3-6} \cmidrule(l){7-9} 
& & \textbf{Classification} & \textbf{VQA} & \textbf{Retrieval} & \textbf{Grounding} & \textbf{IND} & \textbf{OOD} & \textbf{Overall} \\ 
\midrule
\# of Datasets $\rightarrow$ & & 10 & 10 & 12 & 4 & 20 & 16 & 36 \\ 
\midrule

\multicolumn{9}{@{}c}{\textbf{\textit{\footnotesize Zero-shot on MMEB}}} \\ 
\midrule
CLIP (ViT-L)\cite{VLM2Vec} & 0.4B & 42.8 & 9.1 & 53.0 & 51.8 & 37.1 & 38.7 & 39.2 \\
OpenCLIP (ViT-L)\cite{CLIP} & 0.4B & 41.5 & 6.9 & 44.6 & 53.5 & 32.8 & 36.0 & 36.6 \\
Magiclens (ViT-L)\cite{magiclens} & 0.4B & 38.8 & 8.3 & 35.4 & 26.0 & 31.0 & 23.7 & 27.1 \\
SigLIP (So/14)\cite{siglip} & 0.9B & 40.3 & 8.4 & 31.6 & 59.5 & 32.3 & 38.0 & 35.0 \\
BLIP2 (ViT-L)\cite{blip2} & 1.2B & 27.0 & 4.2 & 33.9 & 47.0 & 25.3 & 25.1 & 28.0 \\
CLIP (ViT-BigG/14)\cite{CLIP_bigG14} & 2.5B & 52.3 & 14.0 & 50.5 & 60.3 & 38.9 & 45.8 & 44.3 \\
EVA-CLIP\cite{EVA_CLIP_18B} & 8B & 56.0 & 10.4 & 49.2 & 58.9 & 38.1 & 45.6 & 43.7 \\
\hdashline
E5-V (Phi3.5-V)\cite{E5V} & 4.2B & 39.1 & 9.6 & 38.0 & 57.6 & 33.1 & 31.9 & 36.1 \\
E5-V (LLaVA-1.6)\cite{E5V} & 7B & 39.7 & 10.8 & 39.4 & 60.2 & 34.2 & 33.4 & 37.5 \\
\midrule

\multicolumn{9}{@{}c}{\textbf{\textit{\footnotesize Fine-tuning on MMEB}}} \\ 
\midrule
CLIP (ViT-L)\cite{VLM2Vec} & 0.4B & 55.2 & 19.7 & 53.2 & 62.2 & 47.6 & 42.8 & 47.6 \\
\hdashline
VLM2Vec (Qwen2-VL)\cite{VLM2Vec} & 2B & 59.0 & 49.4 & 65.4 & 73.4 & 66.0 & 52.6 & 60.1 \\
VLM2Vec (Qwen2-VL)\cite{VLM2Vec} & 7B & 62.6 & 57.8 & 69.9 & 81.7 & 72.2 & 57.8 & 65.8 \\
LLaVE (LLaVA-OV)\cite{lan2025llave} & 7B & 65.7 & 65.4 & 70.9 & \textbf{91.9} & \textbf{75.0} & 64.4 & 70.3 \\
QQMM (LLaVA-OV)\cite{qqmm} & 7B & 66.8 & 66.8 & 70.5 & 90.4 & 74.7 & 65.6 & 70.7 \\
UniME (Qwen2-VL)\cite{unime} & 2B & 59.0 & 53.4 & 64.9 & 69.6 & 65.5 & 54.6 & 60.6 \\
UniME (Qwen2-VL)\cite{unime} & 7B & 64.7 & 59.0 & 71.6 & 82.7 & 72.2 & 61.4 & 67.4 \\
UniME (LLaVA-OV)\cite{unime} & 7B & \textbf{66.8} & 66.6 & 70.5 & 90.9 & 74.6 & 65.8 & 70.7 \\
\rowcolor[HTML]{EDEDED}
\mdname (Qwen2-VL) & 2B & 62.1(\textcolor{kcgreen}{+3.1}) & 56.3(\textcolor{kcgreen}{+2.9}) & 68.0(\textcolor{kcgreen}{+3.1}) & 72.7(\textcolor{kcgreen}{+3.1}) & 67.4(\textcolor{kcgreen}{+1.9}) & 58.9(\textcolor{kcgreen}{+4.3}) & 63.6(\textcolor{kcgreen}{+3.0}) \\
\rowcolor[HTML]{EDEDED}
\mdname (Qwen2-VL) & 7B & 64.0(\textcolor{kcred}{-0.7}) & 60.1(\textcolor{kcgreen}{+1.1}) & \textbf{73.1}(\textcolor{kcgreen}{+1.5}) & 82.8(\textcolor{kcgreen}{+0.1}) & 72.0(\textcolor{kcred}{-0.2}) & 63.0(\textcolor{kcgreen}{+1.6}) & 68.0(\textcolor{kcgreen}{+0.6}) \\
\rowcolor[HTML]{EDEDED}
\mdname (LLaVA-OV) & 7B & 65.3(\textcolor{kcred}{-1.5}) & \textbf{67.6}(\textcolor{kcgreen}{+1.0}) & 72.9(\textcolor{kcgreen}{+2.4}) & 90.2(\textcolor{kcred}{-0.7}) & 74.8(\textcolor{kcgreen}{+0.2}) & \textbf{66.7}(\textcolor{kcgreen}{+0.9}) & \textbf{71.2}(\textcolor{kcgreen}{+0.5}) \\
\bottomrule
\end{tabular}
\vspace{-1mm}
\caption{Results on the MMEB benchmark. IND: in-distribution, OOD: out-of-distribution. Scores are average Precision@1. Detailed results are in the supplementary material.}
\vspace{-3mm}
\label{tab:mmeb}
\end{table*}
\subsection{MLLM Judgment Based Training Framework}
\noindent\textbf{\mdname.} 
Previous works~\cite{VLM2Vec,unime} are limited by a rigid one-to-one mapping, which restricts the ability to learn distinctions among diverse negative samples. To address this, as shown in Fig.~\ref{fig:method2}, we propose an MLLM judgment based distribution alignment framework, leveraging soft semantic matching scores as supervised signals to improve representation performance. Specifically, given a query $q$ and its candidate set $\Omega_c = \{c_t, c_1,...,c_k\}$, we input them into the MLLM and extract the last token as embeddings for the query $e_q$ and candidates $\mathrm{E_c} = \{e_c^+, e_{c_1}^-,..,e_{c_k}^-\}$, where $e_c^+$ is the embedding of target candidate and $k$ is the hard negative number for each query. We then compute the relation score matrix between the query embedding $e_q$ and candidate embeddings $\mathrm{E_c}$ as follows:
\begin{equation}
    \small
    \mathbb{P}(e_q, \mathrm{E_c}) = \frac{exp(cos(e_{q}, e_{c}^{+})/\tau)}{exp(cos(e_{q}, e_{c}^{+})/\tau) + \sum_{i=1}^{k}exp(cos(e_{q}, e_{c_{i}^{-}})/\tau)}.
\end{equation}

Based on the semantic matching scores $\mathrm{S_c}=\{s_{q,c_t},s_{q,c_1},...,s_{q,c_k}\}$, we compute the semantic matching score matrix $\mathbb{Q}(\mathrm{S_c})$ derived from the MLLM judgment as follows:
\begin{equation}
    \small
    \mathbb{Q}(\mathrm{S_c}) = \frac{exp(s_{q, c_t}/\tau)}{exp(s_{q, c_t}/\tau) + \sum_{i=1}^{k}exp(s_{q, c_i}/\tau)}.
\end{equation}
To enhance learning robustness and ensure matrix symmetry, we employ JS-Divergence, a symmetric alternative to KL-Divergence~\cite{nielsen2020generalization}. The final loss function $\mathcal{L}$ is defined as:
\begin{equation}
\begin{split}
    \small
    \mathcal{L} = \frac{1}{2} (\frac{1}{N}\sum_{i=1}^{N}	\mathrm{KL}(\mathbb{P}(e_i, \mathrm{E_c}) || \mathbb{Q}(\mathrm{S_c})) + \\  \frac{1}{N}\sum_{i=1}^{N} \mathrm{KL}(\mathbb{Q}(\mathrm{S_c}) || \mathbb{P}(e_i, \mathrm{E_c})) ).
\end{split}
\end{equation}

\noindent{\textbf{\mdrkname.}} Following previous works~\cite{LamRA, mm_embed}, we train a reranking model to enhance retrieval precision following initial embedding-based retrieval. Specifically, we train \mdrkname\ using joint pairwise and listwise approaches to enhance its reranking capability~(refer to Fig.~\ref{fig:method2}). In pairwise training, we construct two pairs for each query $q$ by combining with the positive candidate $c_t$ and the hardest negatives $c_h$. We then instruct \mdrkname\ to output \texttt{YES} for the positive and \texttt{NO} for the negative. The pairwise loss $\mathcal{L}_{pair}$ is computed using the cross-entropy loss function as:
\begin{equation}
\mathcal{L}_{pair} = \mathcal{L}_{ce}(\mathrm{\texttt{YES}}, \eta(q, c_t)) + \mathcal{L}_{ce}(\mathrm{\texttt{NO}}, \eta(q, c_h)),
\end{equation}
where $\eta$ denotes the autoregressive output process of \mdrkname. For listwise training, based on the semantic matching score, we choose top-$x$ candidates~($\{c_1,...c_x\}$) from the hard negative candidates, insert the target candidate $c_{t}$ at a random position and get its index $I_{c_{t}}$. The \mdrkname\ is then prompted to output the position of the ground truth, formulated as:
\begin{equation}
\mathcal{L}_{list} = \mathcal{L}_{ce}(I_{c_{t}}, \eta(q, c_{t}, \{c_1,...c_x\})).
\end{equation}
The final loss function is defined as $\mathcal{L} = \mathcal{L}_{pair} + \mathcal{L}_{list}$. Detailed descriptions of the prompts used for pairwise and listwise training are provided in the supplementary material.

\subsection{Inference Pipeline}
After obtaining \mdname\ and \mdrkname, we integrate them during inference to improve retrieval performance. We initially use \mdname\ embed query and candidate into features and utilize cosine similarity scores to retrieve the top-10 most relevant candidates. Subsequently, \mdrkname\ reranks these candidates based on the following instruction:
\begin{promptblock}
\vspace{-2mm}
\noindent\textit{I will provide you with a query followed by multiple candidates in the format: (1) candidate1 (2) candidate2, etc. Each candidate is independent of the others. Evaluate each candidate against the query, and respond with the number corresponding to the candidate that best meets the requirements of the query. Query:$<$Query$>$, Candidates:$<$Candidate list$>$.}
\vspace{-2mm}
\end{promptblock}

\section{Experiments and Results}

\subsection{Implementation}
We extract query and candidate embeddings using VLM2Vec (Qwen2-VL-7B) to construct a potential hard negative set. We use the Qwen2.5VL-7B to generate the soft semantic matching score. We train \mdname\ using two different multimodal large language models: Qwen2-VL~\cite{wang2024qwen2} and LLaVA-OneVision~\cite{li2024llava}. To optimize GPU memory, we implement LoRA (rank=16) with DeepSpeed ZeRO stage-2~\cite{deepspeed}. The training of \mdname\ is conducted on 8$\times$NVIDIA A800 (80GB) GPUs to accommodate the substantial computational demands. We use 336$\times$336 resolution image inputs, set the accumulated batch size to 1024, with learning rates of 1e-4 (Qwen2-VL) and 2e-5 (LLaVA-OneVision). We set the temperature of the Symmetric KL loss $\tau=0.02$ and sample $k=8$ hard negatives, and train each model for 2,000 steps.

\subsection{Datasets and Evaluation}
\subsubsection{Training Data.} 
Following VLM2Vec~\cite{VLM2Vec} and UniME~\cite{unime}, we employ 20 in-distribution datasets from the MMEB benchmark, which cover four core multimodal tasks: classification, visual question answering, multimodal retrieval, and visual grounding. This comprehensive training corpus, incorporating both unimodal and multimodal input data, totals 662k carefully curated training pairs, ensuring robust model adaptation across diverse multimodal tasks.

\begin{table*}[t!]
\centering
\setlength\tabcolsep{0.75pt}
\renewcommand\arraystretch{1.3}
\fontsize{7pt}{7pt}\selectfont
\begin{tabular}{@{}l c cc cc cc cc ccc@{}}
\toprule
\multicolumn{1}{l}{\multirow{4}{*}{\textbf{Models}}}
& \multicolumn{1}{c}{\multirow{4}{*}{\textbf{\#Parameters}}}
& \multicolumn{4}{c}{\textbf{Short Caption}} 
& \multicolumn{4}{c}{\textbf{Long Caption}} 
& \multicolumn{3}{c}{\textbf{Compositional}} \\
\cmidrule(lr){3-6} \cmidrule(lr){7-10} \cmidrule(l){11-13} 

& & \multicolumn{2}{c}{\textbf{Flickr30K}} & \multicolumn{2}{c}{\textbf{COCO}} 
& \multicolumn{2}{c}{\textbf{ShareGPT4V}} & \multicolumn{2}{c}{\textbf{Urban1K}} 
& \multicolumn{3}{c}{\textbf{SugarCrepe}} \\
\cmidrule(lr){3-4} \cmidrule(lr){5-6} \cmidrule(lr){7-8} \cmidrule(lr){9-10} \cmidrule(l){11-13}

& & $q^t\rightarrow c^i$ & $q^i\rightarrow c^t$ & $q^t\rightarrow c^i$ & $q^i\rightarrow c^t$ 
& $q^t\rightarrow c^i$ & $q^i\rightarrow c^t$ & $q^t\rightarrow c^i$ & $q^i\rightarrow c^t$ 
& Replace & Swap & Add \\
\midrule

OpenCLIP (ViT-L)~\cite{CLIP} & 0.4B & 67.3 & 87.2 & 37.0 & 58.1 & 81.8 & 84.0 & 47.0 & 47.0 & 79.5 & 62.7 & 74.9 \\ 
CLIP (ViT-BigG/14)~\cite{CLIP_bigG14} & 2.5B & 79.5 & 92.9 & 51.3 & 67.3 & 90.1 & 93.6 & 77.8 & 80.7 & 86.5 & 68.9 & 88.4 \\
EVA-CLIP~\cite{EVA_CLIP_18B} & 8B & 80.3 & \textbf{94.5} & 52.0 & 70.1 & 93.1 & 91.2 & 80.4 & 77.8 & 85.9 & 70.3 & 86.7 \\
\hdashline
E5-V (Phi3.5-V)~\cite{E5V} & 4.2B & 72.2 & 79.6 & 44.7 & 53.4 & 86.0 & 88.5 & 83.8 & 83.6 & 88.2 & 66.6 & 75.3 \\
E5-V (LLaVA-1.6)~\cite{E5V} & 7B & 77.3 & 85.7 & 49.1 & 57.6 & 85.1 & 82.1 & 88.9 & 83.2 & 86.3 & 68.7 & 66.9 \\
VLM2Vec (Qwen2-VL)~\cite{VLM2Vec} & 2B & 69.3 & 89.6 & 40.0 & 62.5 & 78.1 & 88.2 & 78.7 & 83.9 & 67.2 & 46.5 & 66.4 \\
VLM2Vec (Qwen2-VL)~\cite{VLM2Vec} & 7B & 80.0 & 94.2 & 49.2 & 68.5 & 78.5 & 90.4 & 94.0 & 94.2 & 70.0 & 51.7 & 72.2 \\
UniME (Qwen2-VL)~\cite{unime} & 2B & 74.9 & 90.6 & 44.0 & 63.5 & 83.6 & 88.6 & 83.3 & 83.2 & 65.6 & 45.2 & 65.7 \\
UniME (Qwen2-VL)~\cite{unime} & 7B & 80.8 & 92.7 & 50.9 & 69.8 & 86.5 & 93.8 & 95.3 & 94.0 & 68.8 & 53.0 & 69.8 \\
UniME (LLaVA-OV)~\cite{unime} & 7B & 83.3 & 94.4 & 54.8 & 74.0 & 93.9 & 89.3 & 94.3 & 95.5 & 80.5 & 65.5 & 82.2 \\
\rowcolor[HTML]{EDEDED}
UniME-V2 (Qwen2-VL) & 2B & 79.8(\textcolor{kcgreen}{+4.9}) & 89.9(\textcolor{kcred}{-0.7}) & 53.7(\textcolor{kcgreen}{+9.7}) & 65.1(\textcolor{kcgreen}{+1.6}) & 91.6(\textcolor{kcgreen}{+8.0}) & 94.2(\textcolor{kcgreen}{+5.6}) & 95.6(\textcolor{kcgreen}{+12.3}) & 92.2(\textcolor{kcgreen}{+9.0}) & 70.9(\textcolor{kcgreen}{+5.3}) & 51.2(\textcolor{kcgreen}{+6.0}) & 70.2(\textcolor{kcgreen}{+4.5}) \\
\rowcolor[HTML]{EDEDED}
UniME-V2 (Qwen2-VL) & 7B & 84.6(\textcolor{kcgreen}{+3.8}) & 93.5(\textcolor{kcgreen}{+0.8}) & 57.3(\textcolor{kcgreen}{+6.4}) & 70.3(\textcolor{kcgreen}{+0.5}) & 94.3(\textcolor{kcgreen}{+0.8}) & \textbf{95.2}(\textcolor{kcgreen}{+1.4}) & \textbf{97.2}(\textcolor{kcgreen}{+1.9}) & 96.3(\textcolor{kcgreen}{+2.3}) & 77.8(\textcolor{kcgreen}{+9.0}) & 62.2(\textcolor{kcgreen}{+9.2}) & 79.0(\textcolor{kcgreen}{+9.2}) \\
\rowcolor[HTML]{EDEDED}
UniME-V2 (LLaVA-OV) & 7B & \textbf{85.5}(\textcolor{kcgreen}{+2.2}) & 93.7(\textcolor{kcred}{-0.7}) & \textbf{60.9}(\textcolor{kcgreen}{+6.1}) & \textbf{74.1}(\textcolor{kcgreen}{+0.1}) & \textbf{95.1}(\textcolor{kcgreen}{+1.2}) & 94.1(\textcolor{kcgreen}{+4.8}) & 96.3(\textcolor{kcgreen}{+2.0}) & \textbf{96.7}(\textcolor{kcgreen}{+1.2}) & \textbf{88.6}(\textcolor{kcgreen}{+8.1}) & \textbf{73.7}(\textcolor{kcgreen}{+8.2}) & \textbf{90.5}(\textcolor{kcgreen}{+8.3}) \\
\bottomrule
\end{tabular}
\vspace{-1mm}
\caption{Zero-shot text-image retrieval results on short caption (Flickr30K, MS-COCO), long caption (ShareGPT4V, Urban1K) and compositional (SugarCrepe) datasets. Scores are Recall@1.}
\vspace{-3mm}
\label{tab:zero-shot-retrieval}
\end{table*} 

\subsubsection{Evaluation.}
In this study, we evaluate \mdname\ across both in-distribution (20 test sets) and out-of-distribution (16 test sets) benchmarks from MMEB~\cite{VLM2Vec} to assess its multimodal embedding capabilities across diverse retrieval tasks. Following standard evaluation protocols~\cite{LamRA, VLM2Vec}, we report Precision, measuring the proportion of correct matches among the top-ranked candidates for each dataset. To further examine the unimodal embedding performance of \mdname, we conduct experiments on multiple cross-modal retrieval tasks, including short-caption image-text retrieval on Flickr30K~\cite{flickr30k} and COCO2014~\cite{MSCOCO2014}, long-caption image-text retrieval on ShareGPT4V~\cite{sharegpt4v} and Urban1K~\cite{longclip}, and compositional retrieval on SugarCrepe~\cite{sugarcrepe}. Consistent with the MMEB benchmark, we use Precision as the primary evaluation metric across all datasets.

\subsection{Main Results}
\subsubsection{Multi-Modal Retrieval.} In Tab.~\ref{tab:mmeb}, we present the performance of the proposed UniME-V2 compared to existing baseline models. Under identical training data and configurations, UniME-V2 consistently achieves notable performance improvements across various foundation models. Specifically, UniME-V2 outperforms VLM2Vec by 3.5\% and 2.2\% on the Qwen2-VL-2B and 7B models, respectively. When built on LLaVA-OneVision as the foundation, UniME-V2 achieves a 0.5\%-0.9\% improvement over previous state-of-the-art models, including QQMM, LLaVE, and UniME. Furthermore, UniME-V2 attains a score of 66.7 on out-of-distribution datasets, significantly exceeding all prior approaches, highlighting its robustness and superior transferability.

\begin{figure}[t!]
\centering
\includegraphics[width=0.95\linewidth]{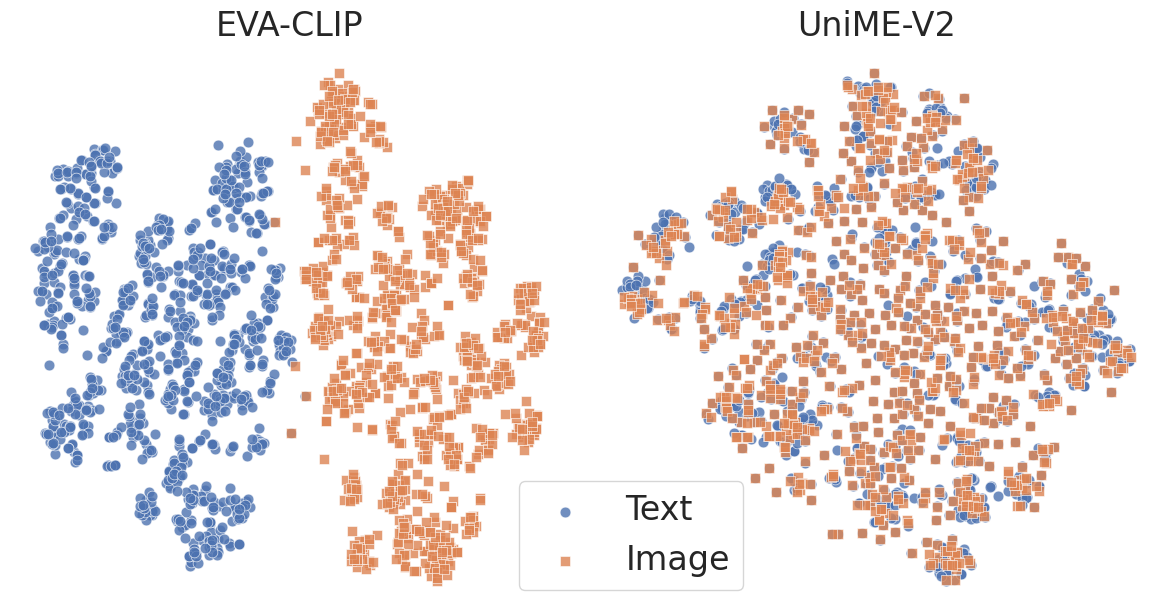}
\vspace{-2mm}
\caption{Comparison of representation distributions between EVA-CLIP-8B and UniME-V2 (LLaVA-OneVision-7B).}
\vspace{-5mm}
\label{fig:distribution}
\end{figure}

\subsubsection{Short \& Long Caption Cross-Modal Retrieval.} We evaluate UniME-V2 on zero-shot cross-modal retrieval tasks. For short-caption datasets, including Flickr30K and MS-COCO, UniME-V2 demonstrates a 2.2\%-9.7\% performance improvement in image-to-text retrieval compared to UniME. In text-to-image retrieval, its performance is comparable to UniME, primarily due to two factors: (1) the limited proportion of text-to-image data in the MMEB training set and (2) the insufficient semantic information in short captions. For long-caption cross-modal retrieval tasks, UniME-V2 achieves significant improvements on ShareGPT4V and Urban1K, benefitting from its enhanced discriminative capability and the richer semantic content provided by detailed captions. Notably, compared to EVA-CLIP-8B, UniME-V2 demonstrates more robust retrieval performance. This is primarily due to its universal multimodal embedding can significantly reduce the modality gap (as shown in Fig.~\ref{fig:distribution}).

\subsubsection{Compositional Cross-Modal Retrieval.} We evaluate the capacity of the UniME-V2 model to discriminate hard negative samples using the compositional benchmark SugarCrepe. As shown in Tab.~\ref{tab:zero-shot-retrieval}, UniME-V2 consistently delivers superior performance across all evaluated metrics. Compared with UniME, our model achieves 5.3\%, 6.0\%, 4.5\% performance improvement using Qwen2-VL-2B. After scaling the model from 2B to 7B, our model also achieves 9.0\%, 9.2\%, and 9.2\% performance improvement. Additionally, UniME-V2 exhibits improvements of 2.7\%, 3.4\%, and 3.8\% compared to EVA-CLIP-8B, underscoring its robust capability to discriminate against hard negative samples.

\begin{table}[t!]
\centering
\small

\setlength\tabcolsep{3pt}
\renewcommand\arraystretch{1.2}
\small
\fontsize{7pt}{7pt}\selectfont
\begin{tabular}{ccccccc}
\toprule
 Embedding Model & Reranker & \#Data & MMEB & $\text{R}_{\text{Short}}$ & $\text{R}_{\text{Long}}$ & $\text{R}_{\text{Compos}}$    \\
 \midrule
 UniME(2B) & --- & --- & 60.6 & 68.3 & 84.7 & 58.8 \\
\mdname(2B)  & --- & --- & 63.6 & 72.1 & 93.4 & 64.1 \\
\mdname(2B) & LamRA(7B) & 1.1M & 67.3 & 76.4 & 96.4 & 87.4 \\
 \rowcolor[HTML]{EDEDED}
\mdname(2B) & \mdname(7B) & 0.6M & \textbf{67.6} & \textbf{76.4} & \textbf{96.9} & \textbf{94.8} \\
\hdashline
UniME(7B) & --- & --- & 67.4 & 73.6 & 92.4 & 63.9 \\
\mdname(7B) & --- & --- & 68.0 & 76.4 & 95.8 & 73.0 \\
\mdname(7B) & LamRA(7B) & 1.1M & 69.1 & 78.3 & 97.2 & 87.4 \\
\rowcolor[HTML]{EDEDED}
\mdname(7B) & \mdname(7B) & 0.6M & \textbf{69.6} & \textbf{78.7} & \textbf{97.5} & \textbf{94.8} \\

\bottomrule
\end{tabular}
\vspace{-2mm}
\caption{Comparison of reranking performance between LamRA and UniME-V2-Reranker using UniME-V2 (Qwen2-VL-7B) and UniME-V2 (Qwen2-VL-2B).}
\vspace{-3mm}
\label{tab:reranking}

\end{table}
\subsubsection{Reranking Comparison.} In Tab.~\ref{tab:reranking}, we compare the performance between LamRA and UniME-V2-Reranker using listwise reranking on the top-5 retrieval results. To ensure fairness, we use the same training parameters and base model (Qwen2.5-VL-7B) as LamRA. When UniME-V2 (Qwen2-VL-2B) is used for retrieval, both LamRA and UniME-V2-Reranker improve performance across four downstream tasks, with UniME-V2-Reranker consistently achieving superior results while utilizing only half the data. Similarly, with UniME-V2 (Qwen2-VL-7B) for retrieval, UniME-V2-Reranker outperforms LamRA, achieving performance gains of 0.5\%, 0.4\%, 0.3\%, and 7.4\% across the four tasks. Notably, UniME-V2-Reranker demonstrates a significant advantage over LamRA in compositional understanding retrieval tasks, attributed to its use of MLLM’s understanding capabilities to extract diverse and high-quality hard samples, which effectively enhance the model’s discriminative power.

\begin{table}[t]
\centering
\setlength\tabcolsep{4pt}
\renewcommand\arraystretch{1.2}
\small
\fontsize{8pt}{8pt}\selectfont
\begin{tabular}{ccccccc}
\toprule
 Hard Negatives & Soft Score & MMEB & $\text{R}_{\text{Short}}$ & $\text{R}_{\text{Long}}$ & $\text{R}_{\text{Compos}}$    \\
 \midrule
 \ding{56} & \ding{56}  & 60.1  & 63.4 & 82.2 & 60.0 \\
 \ding{52} & \ding{56} &  61.6 & 68.9 & 89.8 & 63.7  \\
 \rowcolor[HTML]{EDEDED}
 \ding{52} & \ding{52} & \textbf{63.6} & \textbf{72.1} & \textbf{93.4} & \textbf{64.1}  \\

\bottomrule
\end{tabular}
\vspace{-2mm}
\caption{Ablation study on our proposed MLLM-as-a-Judge hard negatives mining method and MLLM judgment based training framework.}
\vspace{-3mm}
\label{tab:ablation_data_softlabel}
\end{table}

\subsection{Analysis}
\subsubsection{Ablation on Different Components.}
We evaluate the effectiveness of UniME-V2 through ablation studies on the proposed MLLM-as-a-Judge hard negatives mining method and the MLLM judgment based training framework, utilizing Qwen2-VL-2B. As shown in Tab.~\ref{tab:ablation_data_softlabel}, our proposed hard negatives mining method achieves performance improvements of 1.5\%, 5.5\%, 7.6\%, and 3.7\% over direct contrastive learning (e.g., VLM2Vec) on the MMEB, short-retrieval, long-retrieval, and composed-retrieval tasks, respectively. Building on this, the introduction of the MLLM judgment based training framework further enhances the model's discriminative ability by capturing finer semantic distinctions among candidate samples, leading to additional performance gains of 2.0\%, 3.2\%, 3.6\%, and 0.4\% for the corresponding tasks.

\begin{table}[t!]
\centering
\setlength\tabcolsep{6pt}
\renewcommand\arraystretch{1.2}
\small
\fontsize{8pt}{8pt}\selectfont
\begin{tabular}{lccccc}
\toprule
 Judge Model & MMEB & $\text{R}_{\text{Short}}$ & $\text{R}_{\text{Long}}$ & $\text{R}_{\text{Compos}}$    \\
 \midrule
 \rowcolor[HTML]{EDEDED}
 Qwen2.5VL-7B  & \textbf{63.6} & 72.1 & \textbf{93.4} & \textbf{64.1} \\
 InternVL3-8B  & 58.5 & 70.2 & 91.3 & 64.1 \\
 InternVL3-14B  & 63.2 & \textbf{72.9} & 93.1 & 63.2 \\

\bottomrule
\end{tabular}
\vspace{-2mm}
\caption{Ablation study on different MLLM-based judges.}
\vspace{-3mm}
\label{tab:ablation_judges}
\end{table}
\begin{table}[t!]
\centering
\setlength\tabcolsep{7pt}
\renewcommand\arraystretch{1.1}
\small
\fontsize{8pt}{8pt}\selectfont
\begin{tabular}{cccccc}
\toprule
    \#Negatives & MMEB & $\text{R}_{\text{Short}}$ & $\text{R}_{\text{Long}}$ & $\text{R}_{\text{Compos}}$    \\
 \midrule
  4  & 61.3 & 69.2 & 91.0 & 62.4 \\
 6  & 61.8 & 70.8 & 91.7 & 61.2 \\
 \rowcolor[HTML]{EDEDED}
 8  & \textbf{63.6} & \textbf{72.1} & \textbf{93.4} & \textbf{64.1}  \\
 10 & 63.0 & 72.0 & 93.4 & 63.4  \\

\bottomrule
\end{tabular}
\vspace{-2mm}
\caption{Ablation study on the number of hard negatives.}
\vspace{-5mm}
\label{tab:ablation_number}
\end{table}
\subsubsection{Ablation on Different MLLM-based Judges.}
The comprehension ability of the MLLM acting as a judge directly impacts the accuracy of the generated semantic matching scores, thereby influencing the final model performance. Therefore, based on Qwen2-VL-2B, we compare two influential MLLMs in the current open-source community: Qwen2.5-VL-7B, InternVL3-8B, and InternVL3-14B. As shown in Tab.~\ref{tab:ablation_judges}, under the same inference settings, the quality of semantic matching scores produced by Qwen2.5-VL is significantly superior to that of InternVL3-8B, particularly on the MMEB (63.6 v.s. 58.5). When employing InternVL3-14B, there is a notable enhancement in downstream performance compared to Intern3-8B, but it remains slightly inferior to Qwen2.5-7B. The primary reason can be attributed to differences in the distribution of instruction data used during their SFT phase.

\subsubsection{Ablation on the Number of Hard Negatives.}
Tab.~\ref{tab:ablation_number} presents the impact of varying the number of hard negatives based on Qwen2-VL-2B. When the number of hard negative samples increases from 4 to 8, UniME-V2 demonstrates consistent improvements across all evaluation metrics: +2.3\% on MMEB, +2.9\% on short retrieval, +2.4\% on long retrieval, and +1.7\% on composed retrieval. These gains can be attributed to the model's enhanced ability to discriminate between candidates during training. However, further increasing to 10 introduces easier negatives, diminishing discriminative learning, and slightly reducing performance.

\subsubsection{Qualitative Results.}
\begin{figure}[t!]
\centering
\includegraphics[width=\linewidth]{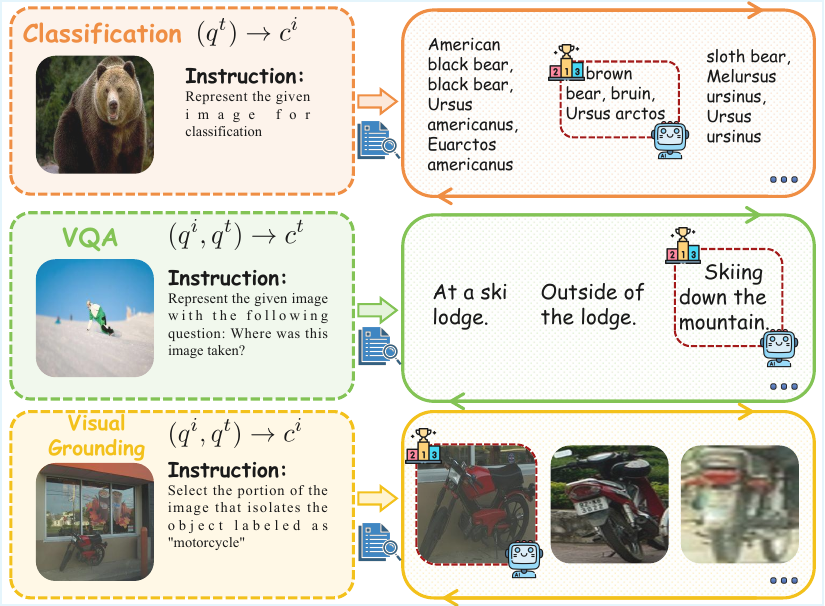}
\vspace{-5mm}
\caption{Qualitative examples. We present the retrieval and reranking results of our method across different tasks.}
\vspace{-5mm}
\label{fig:vis}
\end{figure}

Fig.~\ref{fig:vis} illustrates the qualitative results of our method across various tasks. Retrieval results from UniME-V2 are shown, with the top-1 candidate refined by UniME-V2-Reranker highlighted in red dashed boxes. UniME-V2 effectively retrieves query-relevant candidates, such as ``black bear'' and ``brown bear'' in the first example, while UniME-V2-Reranker further refines the ranking of retrieved results, prioritizing ``brown bear'' over ``black bear''.

\section{Conclusion}
In this paper, we explore how to leverage the advanced understanding capabilities of MLLMs to enhance representation learning and propose a novel Universal Multimodal Embedding model (UniME-V2). Specifically, we first construct a potential hard negative set using global retrieval. We then introduce MLLM-as-a-Judge, which utilizes the robust semantic understanding of MLLMs to assess the alignment of query-candidate pairs and generate soft semantic matching scores. These scores guide hard negative mining by reducing false negative interference and identifying high-quality, diverse hard negatives. Additionally, the scores serve as soft labels, relaxing the rigid one-to-one mapping constraint. By aligning the similarity matrix with the soft semantic matching score matrix, the model learns finer-grained semantic distinctions among candidates, thereby enhancing its discriminative power. To further improve performance, we propose \mdrkname, which incorporates joint pairwise and listwise reranking optimization based on the mined hard negatives. We conduct extensive experiments on the MMEB benchmark and various retrieval tasks and our method achieves state-of-the-art performance on average across all tasks. We hope our work provides insights into universal multimodal representation learning.

\bigskip
\bibliography{aaai2026}

@inproceedings{clip,
  title={Learning transferable visual models from natural language supervision},
  author={Radford, Alec and Kim, Jong Wook and Hallacy, Chris and Ramesh, Aditya and Goh, Gabriel and Agarwal, Sandhini and Sastry, Girish and Askell, Amanda and Mishkin, Pamela and Clark, Jack and others},
  booktitle={ICML},
  year={2021}
}

@inproceedings{longclip,
  title={Long-clip: Unlocking the long-text capability of clip},
  author={Zhang, Beichen and Zhang, Pan and Dong, Xiaoyi and Zang, Yuhang and Wang, Jiaqi},
  booktitle={ECCV},
  year={2024}
}

@article{flame,
  title={FLAME: Frozen Large Language Models Enable Data-Efficient Language-Image Pre-training},
  author={Cao, Anjia and Wei, Xing and Ma, Zhiheng},
  journal={arXiv:2411.11927},
  year={2024}
}

@article{llm2clip,
  title={Llm2clip: Powerful language model unlock richer visual representation},
  author={Huang, Weiquan and Wu, Aoqi and Yang, Yifan and Luo, Xufang and Yang, Yuqing and Hu, Liang and Dai, Qi and Dai, Xiyang and Chen, Dongdong and Luo, Chong and others},
  journal={arXiv:2411.04997},
  year={2024}
}

@article{VLM2Vec,
  title={Vlm2vec: Training vision-language models for massive multimodal embedding tasks},
  author={Jiang, Ziyan and Meng, Rui and Yang, Xinyi and Yavuz, Semih and Zhou, Yingbo and Chen, Wenhu},
  journal={ICLR},
  year={2025}
}

@article{LamRA,
  title={LamRA: Large Multimodal Model as Your Advanced Retrieval Assistant},
  author={Yikun Liu and Pingan Chen and Jiayin Cai and Xiaolong Jiang and Yao Hu and Jiangchao Yao and Yanfeng Wang and Weidi Xie},
  journal={CVPR},
  year={2024}
}

@inproceedings{unime,
   title={Breaking the Modality Barrier: Universal Embedding Learning with Multimodal LLMs},
  author={Gu, Tiancheng and Yang, Kaicheng and Feng, Ziyong and Wang, Xingjun and Zhang, Yanzhao and Long, Dingkun and Chen, Yingda and Cai, Weidong and Deng, Jiankang},
  booktitle={ACM MM},
  year={2025}
}

@inproceedings{degla,
  title={Decoupled Global-Local Alignment for Improving Compositional Understanding},
  author={Hu, Xiaoxing and Yang, Kaicheng and Wang, Jun and Xu, Haoran and Feng, Ziyong and Wang, Yupei},
  booktitle={ACM MM},
  year={2025}
}

@article{negclip,
  title={When and why vision-language models behave like bags-of-words, and what to do about it?},
  author={Yuksekgonul, Mert and Bianchi, Federico and Kalluri, Pratyusha and Jurafsky, Dan and Zou, James},
  journal={arXiv:2210.01936},
  year={2022}
}

@article{tschannen2023image,
  title={Image captioners are scalable vision learners too},
  author={Tschannen, Michael and Kumar, Manoj and Steiner, Andreas and Zhai, Xiaohua and Houlsby, Neil and Beyer, Lucas},
  journal={NeurIPS},
  year={2023}
}

@inproceedings{zheng2025gradient,
  title={Gradient-Attention Guided Dual-Masking Synergetic Framework for Robust Text-based Person Retrieval},
  author={Zheng, Tianlu and Zhang, Yifan and An, Xiang and Feng, Ziyong and Yang, Kaicheng and Ding, Qichuan},
  booktitle={EMNLP},
  year={2025}
}

@article{xue2025improve,
  title={Improve Multi-Modal Embedding Learning via Explicit Hard Negative Gradient Amplifying},
  author={Xue, Youze and Li, Dian and Liu, Gang},
  journal={arXiv preprint arXiv:2506.02020},
  year={2025}
}

@article{NVembed,
  title={Nv-embed: Improved techniques for training llms as generalist embedding models},
  author={Lee, Chankyu and Roy, Rajarshi and Xu, Mengyao and Raiman, Jonathan and Shoeybi, Mohammad and Catanzaro, Bryan and Ping, Wei},
  journal={ICLR},
  year={2024}
}

@article{LLM2Vec,
  title={Llm2vec: Large language models are secretly powerful text encoders},
  author={BehnamGhader, Parishad and Adlakha, Vaibhav and Mosbach, Marius and Bahdanau, Dzmitry and Chapados, Nicolas and Reddy, Siva},
  journal={COLM},
  year={2024}
}

@article{mteb,
  author = {Muennighoff, Niklas and Tazi, Nouamane and Magne, Lo{\"\i}c and Reimers, Nils},
  title = {MTEB: Massive Text Embedding Benchmark},
  journal={arXiv:2210.07316},
  year = {2022}
}

@article{E5V,
  title={E5-v: Universal embeddings with multimodal large language models},
  author={Jiang, Ting and Song, Minghui and Zhang, Zihan and Huang, Haizhen and Deng, Weiwei and Sun, Feng and Zhang, Qi and Wang, Deqing and Zhuang, Fuzhen},
  journal={arXiv:2407.12580},
  year={2024}
}

@inproceedings{cc3m,
  title={Conceptual 12m: Pushing web-scale image-text pre-training to recognize long-tail visual concepts},
  author={Changpinyo, Soravit and Sharma, Piyush and Ding, Nan and Soricut, Radu},
  booktitle={CVPR},
  pages={3558--3568},
  year={2021}
}

@article{wang2024qwen2,
  title={Qwen2-vl: Enhancing vision-language model's perception of the world at any resolution},
  author={Wang, Peng and Bai, Shuai and Tan, Sinan and Wang, Shijie and Fan, Zhihao and Bai, Jinze and Chen, Keqin and Liu, Xuejing and Wang, Jialin and Ge, Wenbin and others},
  journal={arXiv preprint arXiv:2409.12191},
  year={2024}
}

@article{li2024llava,
  title={Llava-onevision: Easy visual task transfer},
  author={Li, Bo and Zhang, Yuanhan and Guo, Dong and Zhang, Renrui and Li, Feng and Zhang, Hao and Zhang, Kaichen and Zhang, Peiyuan and Li, Yanwei and Liu, Ziwei and others},
  journal={arXiv preprint arXiv:2408.03326},
  year={2024}
}

@inproceedings{gu2025realsyn,
  title={RealSyn: An Effective and Scalable Multimodal Interleaved Document Transformation Paradigm},
  author={Gu, Tiancheng and Yang, Kaicheng and Zhang, Chaoyi and Xie, Yin and An, Xiang and Feng, Ziyong and Liu, Dongnan and Cai, Weidong and Deng, Jiankang},
  booktitle={ACM MM},
  year={2025}
}

@inproceedings{vila,
  title={Vila: On pre-training for visual language models},
  author={Lin, Ji and Yin, Hongxu and Ping, Wei and Molchanov, Pavlo and Shoeybi, Mohammad and Han, Song},
  booktitle={CVPR},
  year={2024}
}

@article{kosmos,
  title={Kosmos-2: Grounding multimodal large language models to the world},
  author={Peng, Zhiliang and Wang, Wenhui and Dong, Li and Hao, Yaru and Huang, Shaohan and Ma, Shuming and Wei, Furu},
  journal={arXiv:2306.14824},
  year={2023}
}

@article{LLaVA,
  title={Visual instruction tuning},
  author={Liu, Haotian and Li, Chunyuan and Wu, Qingyang and Lee, Yong Jae},
  journal={NeurIPS},
  year={2023}
}

@misc{wang2023cogvlm,
      title={CogVLM: Visual Expert for Pretrained Language Models}, 
      author={Weihan Wang and Qingsong Lv and Wenmeng Yu and Wenyi Hong and Ji Qi and Yan Wang and Junhui Ji and Zhuoyi Yang and Lei Zhao and Xixuan Song and Jiazheng Xu and Bin Xu and Juanzi Li and Yuxiao Dong and Ming Ding and Jie Tang},
      year={2023},
      eprint={2311.03079},
      archivePrefix={arXiv},
      primaryClass={cs.CV}
}

@article{xie2024croc,
  title={Croc: Pretraining large multimodal models with cross-modal comprehension},
  author={Xie, Yin and Yang, Kaicheng and Yang, Ninghua and Deng, Weimo and Dai, Xiangzi and Gu, Tiancheng and Wang, Yumeng and An, Xiang and Zhao, Yongle and Feng, Ziyong and others},
  journal={arXiv preprint arXiv:2410.14332},
  year={2024}
}

@article{bai2025qwen2,
  title={Qwen2. 5-vl technical report},
  author={Bai, Shuai and Chen, Keqin and Liu, Xuejing and Wang, Jialin and Ge, Wenbin and Song, Sibo and Dang, Kai and Wang, Peng and Wang, Shijie and Tang, Jun and others},
  journal={arXiv preprint arXiv:2502.13923},
  year={2025}
}

@article{mm_embed,
  title={Mm-embed: Universal multimodal retrieval with multimodal llms},
  author={Lin, Sheng-Chieh and Lee, Chankyu and Shoeybi, Mohammad and Lin, Jimmy and Catanzaro, Bryan and Ping, Wei},
  journal={arXiv preprint arXiv:2411.02571},
  year={2024}
}

@inproceedings{flickr30k,
  title={Flickr30k entities: Collecting region-to-phrase correspondences for richer image-to-sentence models},
  author={Plummer, Bryan A and Wang, Liwei and Cervantes, Chris M and Caicedo, Juan C and Hockenmaier, Julia and Lazebnik, Svetlana},
  booktitle={ICCV},
  year={2015}
}

@inproceedings{MSCOCO2014,
  title={Microsoft coco: Common objects in context},
  author={Lin, Tsung-Yi and Maire, Michael and Belongie, Serge and Hays, James and Perona, Pietro and Ramanan, Deva and Doll{\'a}r, Piotr and Zitnick, C Lawrence},
  booktitle={ECCV},
  year={2014},
}

@article{sugarcrepe,
  title={Sugarcrepe: Fixing hackable benchmarks for vision-language compositionality},
  author={Hsieh, Cheng-Yu and Zhang, Jieyu and Ma, Zixian and Kembhavi, Aniruddha and Krishna, Ranjay},
  journal={NeurIPS},
  year={2023}
}

@inproceedings{sharegpt4v,
  title={Sharegpt4v: Improving large multi-modal models with better captions},
  author={Chen, Lin and Li, Jinsong and Dong, Xiaoyi and Zhang, Pan and He, Conghui and Wang, Jiaqi and Zhao, Feng and Lin, Dahua},
  booktitle={ECCV},
  year={2024}
}

@inproceedings{deepspeed,
  title={Deepspeed-inference: enabling efficient inference of transformer models at unprecedented scale},
  author={Aminabadi, Reza Yazdani and Rajbhandari, Samyam and Awan, Ammar Ahmad and Li, Cheng and Li, Du and Zheng, Elton and Ruwase, Olatunji and Smith, Shaden and Zhang, Minjia and Rasley, Jeff and others},
  booktitle={SC22: International Conference for High Performance Computing, Networking, Storage and Analysis},
  pages={1--15},
  year={2022},
  organization={IEEE}
}

@article{magiclens,
  title={Magiclens: Self-supervised image retrieval with open-ended instructions},
  author={Zhang, Kai and Luan, Yi and Hu, Hexiang and Lee, Kenton and Qiao, Siyuan and Chen, Wenhu and Su, Yu and Chang, Ming-Wei},
  journal={arXiv:2403.19651},
  year={2024}
}

@inproceedings{blip2,
  title={Blip-2: Bootstrapping language-image pre-training with frozen image encoders and large language models},
  author={Li, Junnan and Li, Dongxu and Savarese, Silvio and Hoi, Steven},
  booktitle={ICML},
  year={2023}
}

@inproceedings{siglip,
  title={Sigmoid loss for language image pre-training},
  author={Zhai, Xiaohua and Mustafa, Basil and Kolesnikov, Alexander and Beyer, Lucas},
  booktitle={ICCV},
  year={2023}
}

@article{CLIP_bigG14,
  title={Reproducible scaling laws for contrastive language-image learning},
  author={Cherti, Mehdi and Beaumont, Romain and Wightman, Ross and Wortsman, Mitchell and Ilharco, Gabriel and Gordon, Cade and Schuhmann, Christoph and Schmidt, Ludwig and Jitsev, Jenia},
  journal={arXiv:2212.07143},
  year={2022}
}

@article{EVA_CLIP_18B,
  title={EVA-CLIP-18B: Scaling CLIP to 18 Billion Parameters}, 
  author={Quan Sun and Jinsheng Wang and Qiying Yu and Yufeng Cui and Fan Zhang and Xiaosong Zhang and Xinlong Wang},
  journal={arXiv:2402.04252},
  year={2023}
}

@inproceedings{chen2024mllm,
  title={Mllm-as-a-judge: Assessing multimodal llm-as-a-judge with vision-language benchmark},
  author={Chen, Dongping and Chen, Ruoxi and Zhang, Shilin and Wang, Yaochen and Liu, Yinuo and Zhou, Huichi and Zhang, Qihui and Wan, Yao and Zhou, Pan and Sun, Lichao},
  booktitle={ICML},
  year={2024}
}

@article{zheng2023judging,
  title={Judging llm-as-a-judge with mt-bench and chatbot arena},
  author={Zheng, Lianmin and Chiang, Wei-Lin and Sheng, Ying and Zhuang, Siyuan and Wu, Zhanghao and Zhuang, Yonghao and Lin, Zi and Li, Zhuohan and Li, Dacheng and Xing, Eric and others},
  journal={NeuriPS},
  volume={36},
  pages={46595--46623},
  year={2023}
}

@article{lan2025llave,
  title={Llave: Large language and vision embedding models with hardness-weighted contrastive learning},
  author={Lan, Zhibin and Niu, Liqiang and Meng, Fandong and Zhou, Jie and Su, Jinsong},
  journal={arXiv preprint arXiv:2503.04812},
  year={2025}
}

@article{qqmm,
  title={Improve Multi-Modal Embedding Learning via Explicit Hard Negative Gradient Amplifying},
  author={Xue, Youze and Li, Dian and Liu, Gang},
  journal={arXiv preprint arXiv:2506.02020},
  year={2025}
}

@inproceedings{hamza2025llava,
  title={Llava needs more knowledge: Retrieval augmented natural language generation with knowledge graph for explaining thoracic pathologies},
  author={Hamza, Ameer and Ahn, Yong Hyun and Lee, Sungyoung and Kim, Seong Tae and others},
  booktitle={AAAI},
  year={2025}
}

@inproceedings{dong2024toward,
  title={Toward general instruction-following alignment for retrieval-augmented generation},
  author={Dong, Guanting and Song, Xiaoshuai and Zhu, Yutao and Qiao, Runqi and Dou, Zhicheng and Wen, Ji-Rong},
  booktitle={AAAI},
  year={2025}
}

@article{nielsen2020generalization,
  title={On a generalization of the Jensen--Shannon divergence and the Jensen--Shannon centroid},
  author={Nielsen, Frank},
  journal={Entropy},
  year={2020}
}

@article{gao2021scaling,
  title={Scaling deep contrastive learning batch size under memory limited setup},
  author={Gao, Luyu and Zhang, Yunyi and Han, Jiawei and Callan, Jamie},
  journal={arXiv preprint arXiv:2101.06983},
  year={2021}
}

@article{wang2025vl,
  title={Vl-rethinker: Incentivizing self-reflection of vision-language models with reinforcement learning},
  author={Wang, Haozhe and Qu, Chao and Huang, Zuming and Chu, Wei and Lin, Fangzhen and Chen, Wenhu},
  journal={arXiv preprint arXiv:2504.08837},
  year={2025}
}

@article{wang2025code,
  title={To code or not to code? adaptive tool integration for math language models via expectation-maximization},
  author={Wang, Haozhe and Li, Long and Qu, Chao and Zhu, Fengming and Xu, Weidi and Chu, Wei and Lin, Fangzhen},
  journal={arXiv preprint arXiv:2502.00691},
  year={2025}
}

@article{an2025unictokens,
  title={UniCTokens: Boosting Personalized Understanding and Generation via Unified Concept Tokens},
  author={An, Ruichuan and Yang, Sihan and Zhang, Renrui and Shen, Zijun and Lu, Ming and Dai, Gaole and Liang, Hao and Guo, Ziyu and Yan, Shilin and Luo, Yulin and others},
  journal={arXiv preprint arXiv:2505.14671},
  year={2025}
}

@article{an2024mc,
  title={Mc-llava: Multi-concept personalized vision-language model},
  author={An, Ruichuan and Yang, Sihan and Lu, Ming and Zhang, Renrui and Zeng, Kai and Luo, Yulin and Cao, Jiajun and Liang, Hao and Chen, Ying and She, Qi and others},
  journal={arXiv preprint arXiv:2411.11706},
  year={2024}
}

@inproceedings{
li2025miv,
title={M{\texttwosuperior}{IV}: Towards Efficient and Fine-grained Multimodal In-Context Learning via Representation Engineering},
author={Yanshu Li and Yi Cao and Hongyang He and Qisen Cheng and Xiang Fu and Xi Xiao and Tianyang Wang and Ruixiang Tang},
booktitle={Second Conference on Language Modeling},
year={2025},
url={https://openreview.net/forum?id=9ffYcEiNw9}
}

@inproceedings{yang2025clip,
  title={Clip-cid: Efficient clip distillation via cluster-instance discrimination},
  author={Yang, Kaicheng and Gu, Tiancheng and An, Xiang and Jiang, Haiqiang and Dai, Xiangzi and Feng, Ziyong and Cai, Weidong and Deng, Jiankang},
  booktitle={AAAI},
  volume={39},
  number={20},
  pages={21974--21982},
  year={2025}
}

@article{schuhmann2021laion,
  title={Laion-400m: Open dataset of clip-filtered 400 million image-text pairs},
  author={Schuhmann, Christoph and Vencu, Richard and Beaumont, Romain and Kaczmarczyk, Robert and Mullis, Clayton and Katta, Aarush and Coombes, Theo and Jitsev, Jenia and Komatsuzaki, Aran},
  journal={arXiv preprint arXiv:2111.02114},
  year={2021}
}

@inproceedings{gu2024rwkv,
  title={Rwkv-clip: A robust vision-language representation learner},
  author={Gu, Tiancheng and Yang, Kaicheng and An, Xiang and Feng, Ziyong and Liu, Dongnan and Cai, Weidong and Deng, Jiankang},
  booktitle={EMNLP},
  year={2024}
}

@article{an2025llava,
  title={LLaVA-OneVision-1.5: Fully Open Framework for Democratized Multimodal Training},
  author={An, Xiang and Xie, Yin and Yang, Kaicheng and Zhang, Wenkang and Zhao, Xiuwei and Cheng, Zheng and Wang, Yirui and Xu, Songcen and Chen, Changrui and Wu, Chunsheng and others},
  journal={arXiv preprint arXiv:2509.23661},
  year={2025}
}

@inproceedings{tang2025seeing,
  title={Seeing Far and Clearly: Mitigating Hallucinations in MLLMs with Attention Causal Decoding},
  author={Tang, Feilong and Liu, Chengzhi and Xu, Zhongxing and Hu, Ming and Huang, Zile and Xue, Haochen and Chen, Ziyang and Peng, Zelin and Yang, Zhiwei and Zhou, Sijin and others},
  booktitle={CVPR},
  pages={26147--26159},
  year={2025}
}

@inproceedings{tang2025intervening,
  title={Intervening anchor token: Decoding strategy in alleviating hallucinations for MLLMs},
  author={Tang, Feilong and Huang, Zile and Liu, Chengzhi and Sun, Qiang and Yang, Harry and Lim, Ser-Nam},
  booktitle={ICLR},
  year={2025}
}

@inproceedings{lei2023mcad,
  title={MCAD: Multi-teacher Cross-modal Alignment Distillation for efficient image-text retrieval},
  author={Lei, Youbo and He, Feifei and Chen, Chen and Mo, Yingbin and Li, Si Jia and Xie, Defeng and Lu, Haonan},
  booktitle={NAACL},
  year={2024}
}

@inproceedings{huang2024cross,
  title={Cross-modal and uni-modal soft-label alignment for image-text retrieval},
  author={Huang, Hailang and Nie, Zhijie and Wang, Ziqiao and Shang, Ziyu},
  booktitle={AAAI},
  year={2024}
}

@inproceedings{andonian2022robust,
  title={Robust cross-modal representation learning with progressive self-distillation},
  author={Andonian, Alex and Chen, Shixing and Hamid, Raffay},
  booktitle={CVPR},
  year={2022}
}
\clearpage
This supplementary material elaborates on our experimental setup, covering training configurations, instruction prompts for \mdrkname\, and evaluation benchmarks for retrieval tasks. It also includes extended results such as an ablation study on temperature and a detailed performance analysis on the MMEB benchmark. Additionally, we provide supplementary visualizations of training data samples, retrieval outputs, and reranking results.

\section{Detail Experiment Setting}
\label{sec: experiment setting}

\subsection{Training Details}
We provide the training configurations of \mdname\ in Tab.~\ref{tab:supp_training_parameters_embed} and \mdrkname\ in Tab.~\ref{tab:supp_training_parameters_rerank}.

\noindent\textbf{\mdname: } Following UniME~\cite{unime}, we adopt identical experimental settings for training \mdname. We configure LoRA rank as 16 and employ GradCache~\cite{gao2021scaling} for efficient training over 2,000 steps using 8$\times$A800 GPUs (80GB memory each). The learning rate is set to 1$\times$10$^{-4}$ for the Qwen series and 2$\times$10$^{-5}$ for LLaVA-OneVision. Due to memory constraints, the input resolution is fixed at 336 for LLaVA-OneVision and 672 for the Qwen series.

\begin{table}[h!]
\centering
\small
\label{tab:training_details}
\begin{tabular}{@{}lcc@{}}
\toprule
\multicolumn{1}{c}{\textbf{Hyperparameter}} & \textbf{Qwen2-VL-2B/7B} & \textbf{LLaVA-OV-7B} \\
\midrule
Training samples & \multicolumn{2}{c}{662K} \\
Batch size & \multicolumn{2}{c}{1024} \\
Learning rate & 1$\times$10$^{-4}$ & 2$\times$10$^{-5}$ \\
LoRA rank & \multicolumn{2}{c}{16} \\
Training steps & \multicolumn{2}{c}{2000} \\
Optimizer & \multicolumn{2}{c}{AdamW} \\
Infra & \multicolumn{2}{c}{GradCache} \\
Max length & \multicolumn{2}{c}{4096} \\
temperature & \multicolumn{2}{c}{0.02} \\
\#Hard Negatives & \multicolumn{2}{c}{8} \\
Image Resolution & 672 & 336 \\
Precision & \multicolumn{2}{c}{BF16} \\
GPU configuration & \multicolumn{2}{c}{8$\times$A800} \\
Random Seed & \multicolumn{2}{c}{42} \\
\bottomrule
\end{tabular}
\caption{Training hyperparameters and computational requirements for \mdname\ (Qwen2-VL-2B/7B) and \mdname\ (LLaVA-OneVision-7B).}
\label{tab:supp_training_parameters_embed}
\vspace{-3mm}
\end{table}

\noindent\textbf{\mdrkname: } Following LamRA's experimental setup~\cite{LamRA}, we adopt Qwen2.5-VL-7B as the backbone for \mdrkname. The model is trained using LoRA with a rank of 128 for 1 epoch almost 2,000 steps. All experiments are conducted using the lmms-finetune infrastructure with a maximum sequence length of 4096 tokens.

\begin{table}[t!]
\centering
\begin{tabular}{@{}lc@{}}
\toprule
\multicolumn{1}{c}{\textbf{Hyperparameter}} & \textbf{Qwen2.5-VL-7B} \\
\midrule
Training samples & \multicolumn{1}{c}{662K} \\
Batch size & \multicolumn{1}{c}{64} \\
Learning rate & \multicolumn{1}{c}{2$\times$10$^{-5}$} \\
LoRA rank & \multicolumn{1}{c}{128} \\
Training epochs & \multicolumn{1}{c}{1} \\
Optimizer & \multicolumn{1}{c}{AdamW} \\
Infra & \multicolumn{1}{c}{lmms-finetune} \\
Max length & \multicolumn{1}{c}{4096} \\
Precision & \multicolumn{1}{c}{BF16} \\
DeepSpeed Stage & \multicolumn{1}{c}{2} \\
GPU configuration & \multicolumn{1}{c}{8$\times$A800} \\
Random Seed & \multicolumn{1}{c}{42} \\
\bottomrule
\end{tabular}
\caption{Training hyperparameters and computational requirements for \mdrkname\ (Qwen2.5-VL-7B).}
\label{tab:supp_training_parameters_rerank}
\end{table}

\subsection{Detail Instruction Prompt for \mdrkname} 
The prompt template employed for \textbf{pairwise} training of \mdrkname\ is presented below:
\begin{promptblock}
\noindent\textit{I will provide you with a query and a candidate. Please evaluate whether the candidate meets the requirements of the query. If it does, respond with 'Yes'; if it doesn't, respond with 'No'. Query:$<$Query$>$, Candidate:$<$Candidate$>$.}
\end{promptblock}

The prompt used for \textbf{listwise} training of \mdrkname\ is shown below:
\begin{promptblock}
\noindent\textit{I will provide you with a query followed by multiple candidates in the format: (1) cand1 (2) cand2, etc. Each candidate is independent of the others. Evaluate each candidate against the query, and respond with the number corresponding to the candidate that best meets the requirements of the query. Query:$<$Query$>$, Candidates:$<$Candidate list$>$.}
\end{promptblock}

\subsection{Retrieval Task Evaluation Benchmarks} 
We evaluate \mdname\ and \mdrkname\ on diverse retrieval benchmarks, including short-caption, long-caption, and compositional image-text tasks (Tab.~\ref{tab:supp_evaluation_benchmark}). For each benchmark, we follow the standard evaluation protocol. In retrieval tasks, we primarily report Recall@1 as the evaluation metric, using the prompt ``\textit{Represent the image/text}" for both image and text instructions.

\begin{table}[h!]
\centering
\footnotesize 
\setlength{\tabcolsep}{0.8mm} 
\begin{tabular}{lccc}
    \toprule
    \textbf{Benchmark} & \textbf{Zero-shot} & \textbf{\#Queries} & \textbf{\#Cands} \\
    \midrule
    Flickr30K~\cite{flickr30k} & \ding{52} & 1K & 5K \\
    COCO~\cite{MSCOCO2014} & \ding{52} & 5K & 25K \\
    ShareGPT4V~\cite{sharegpt4v} & \ding{52} & 1K & 1K\\
    Urban1K~\cite{longclip} & \ding{52} & 1K & 1K\\
    SugarCrepe~\cite{sugarcrepe} & \ding{52} & 7.5K & 2 \\
    \bottomrule
\end{tabular}
\caption{Summary of the evaluation benchmarks. \# Queries represents the number of test queries, and \# Cands denotes the number of test candidates per query.}
\label{tab:supp_evaluation_benchmark}
\end{table}

\section{External Results}
\label{sec: external results}

\subsection{Ablation on the Temperature}

We conduct additional experiments with \mdname\ (Qwen2-VL-2B) to analyze the impact of temperature in the final loss function. As evidenced by Tab.~\ref{tab:supp_ablation_temperature}, a temperature value of 0.02 yields optimal performance across all evaluation metrics, including MMEB, short \& long retrieval, and compositional retrieval tasks.

\begin{table}[t!]
\centering
\setlength\tabcolsep{7pt}
\renewcommand\arraystretch{1.1}
\small
\fontsize{8pt}{8pt}\selectfont
\begin{tabular}{cccccc}
\toprule
    Temperature & MMEB & $\text{R}_{\text{Short}}$ & $\text{R}_{\text{Long}}$ & $\text{R}_{\text{Compos}}$    \\
 \midrule
 0.03  & 61.9 & 70.6 & 92.0 & 65.8 \\
\rowcolor[HTML]{EDEDED} 
 0.02  & \textbf{63.6} & \textbf{72.1} & \textbf{93.4} & \textbf{64.1}  \\
 0.01  & 62.1 & 70.1 & 91.1 & 66.5  \\

\bottomrule
\end{tabular}
\caption{Ablation study on the temperature. We report the mean scores on the MMEB benchmark, short and long cross-modal retrieval, as well as compositional cross-modal retrieval.}
\label{tab:supp_ablation_temperature}
\end{table}

\subsection{Specific Results on the MMEB Benchmark}
\begin{table*}[]
\centering
\resizebox{0.9\textwidth}{!}{
\begin{tabular}{lcccccccc}
\toprule
 \rowcolor{gray!30} & \textbf{BLIP2} & \textbf{MagicLens} & \textbf{EVA-CLIP} & \textbf{E5-V} & \textbf{VLM2Vec} & \textbf{UniME} & \textbf{\mdname} & \textbf{\mdname$^\dag$} \\
\midrule
\rowcolor{orange!30} \textbf{Classification (10 tasks)} & & & & & & & & \\
ImageNet-1K          & 10.3 & 48.0 & 75.0 & 40.5 & 66.5 & 71.3 & 80.3 & 78.8 \\
N24News              & 36.0 & 33.7 & 33.8 & 31.5 & 76.4 & 79.5 & 66.9 & 66.6 \\
HatefulMemes         & 49.6 & 49.0 & 49.3 & 49.3 & 60.9 & 64.6 & 65.9 & 65.3 \\
VOC2007              & 52.1 & 51.6 & 44.3 & 76.7 & 84.0 & 90.4 & 84.9 & 92.0 \\
SUN397               & 34.5 & 57.0 & 62.7 & 52.3 & 73.2 & 75.9 & 78.9 & 78.7 \\
\rowcolor{yellow!15} Place365            & 21.5 & 31.5 & 38.7 & 32.0 & 42.1 & 45.6 & 42.5 & 42.9 \\
\rowcolor{yellow!15} ImageNet-A          & 3.2  & 8.0  & 54.8 & 18.2 & 39.9 & 45.5 & 53.7 & 48.0 \\
\rowcolor{yellow!15} ImageNet-R          & 39.7 & 70.9 & 95.4 & 56.7 & 74.6 & 78.4 & 87.9 & 89.3 \\
\rowcolor{yellow!15} ObjectNet           & 20.6 & 31.6 & 67.8 & 34.2 & 34.3 & 36.4 & 35.0 & 73.1 \\
\rowcolor{yellow!15} Country-211         & 2.5  & 6.2  & 38.7 & 5.9 & 16.1 & 18.7 & 32.3 & 19.8 \\
\textit{All Classification} & 27.0 & 38.8 & 56.0 & 39.7 & 56.8 & 60.6 & 64.0 & 65.3\\
\midrule

\rowcolor{blue!30} \textbf{VQA (10 tasks)} & & & & & & & & \\
OK-VQA               & 8.7  & 12.7 & 9.9 & 15.1 & 66.5 & 68.3 & 59.3 & 71.9 \\
A-OKVQA              & 3.2  & 2.9  & 2.8 & 4.7 & 54.9 & 58.7 & 32.3 & 71.4 \\
DocVQA               & 2.6  & 3.0  & 7.4 & 9.1 & 64.4 & 67.6 & 91.2 & 92.6 \\
InfographicsVQA      & 2.0  & 5.9  & 6.0 & 8.7 & 34.8 & 37.0 & 63.9 & 63.5 \\
ChartQA              & 0.5  & 0.9  & 1.5 & 4.2 & 33.1 & 33.4 & 56.9 & 55.8 \\
Visual7W             & 1.3  & 2.5  & 2.2 & 4.5 & 49.8 & 51.7 & 60.1 & 62.5 \\
\rowcolor{yellow!15} ScienceQA           & 6.8  & 5.2  & 14.1 & 9.6 & 37.3 & 40.5 & 44.5 & 54.0 \\
\rowcolor{yellow!15} VizWiz              & 4.0  & 1.7  & 4.3 & 8.6 & 39.9 & 42.7 & 47.4 & 53.7 \\
\rowcolor{yellow!15} GQA                 & 9.7  & 43.5 & 44.7 & 34.1 & 57.3 & 63.6 & 55.8 & 69.5 \\
\rowcolor{yellow!15} TextVQA             & 3.3  & 4.6  & 10.8 & 9.5 & 65.7 & 65.2 & 78.4 & 84.5 \\
\textit{All VQA}      & 4.2  & 8.3  & 10.4 & 10.8 & 50.4 & 52.9 & 60.1 & 67.6 \\
\midrule

\rowcolor{green!30} \textbf{Retrieval (12 tasks)} & & & & & & & & \\
VisDial              & 18.0 & 24.8 & 20.4 & 57.6 & 75.3 & 79.7 & 83.4 & 84.2 \\
CIRR                 & 9.8  & 39.1 & 36.0 & 41.0 & 51.3 & 52.2 & 64.0 & 65.5 \\
VisualNews\_t2i      & 48.1 & 50.7 & 82.4 & 43.9 & 70.7 & 74.8 & 79.9 & 77.3 \\
VisualNews\_i2t      & 13.5 & 21.1 & 88.2 & 46.8 & 75.2 & 78.8 & 83.5 & 79.2 \\
MSCOCO\_t2i          & 53.7 & 54.1 & 65.3 & 68.6 & 69.9 & 74.9 & 77.7 & 79.1 \\
MSCOCO\_i2t          & 20.3 & 40.0 & 67.2 & 54.8 & 67.7 & 73.8 & 73.0 & 75.2 \\
NIGHTS               & 56.5 & 58.1 & 0.2 & 0.1 & 63.3 & 66.2 & 69.3 & 68.1 \\
WebQA                & 55.4 & 43.0 & 70.9 & 33.7 & 83.6 & 89.8 & 91.5 & 90.6 \\
\rowcolor{yellow!15} FashionIQ           & 9.3  & 11.2 & 16.1 & 11.2 & 15.2 & 16.5 & 28.5 & 26.4 \\
\rowcolor{yellow!15} Wiki-SS-NQ          & 28.7 & 18.7 & 46.7 & 61.0 & 63.4 & 66.6 & 68.8 & 71.2 \\
\rowcolor{yellow!15} OVEN                & 39.5 & 1.6  & 1.8 & 0.5 & 49.6 & 55.7 & 71.2 & 68.0 \\
\rowcolor{yellow!15} EDIS                & 54.4 & 62.6 & 95.6 & 53.8 & 73.7 & 86.2 & 84.4 & 88.2 \\
\textit{All Retrieval} & 33.9 & 35.4 & 49.2 & 39.4 & 63.3 & 67.9 & 73.1 & 72.9 \\
\midrule

\rowcolor{purple!30} \textbf{Visual Grounding (4 tasks)} & & & & & & & & \\
MSCOCO               & 28.9 & 22.1 & 35.8 & 41.7 & 77.0 & 76.5 & 69.3 & 78.2 \\
\rowcolor{yellow!15} RefCOCO             & 47.4 & 22.8 & 59.9 & 62.2 & 85.9 & 89.3 & 88.4 & 94.6 \\
\rowcolor{yellow!15} RefCOCO-matching    & 59.5 & 35.6 & 70.0 & 74.9 & 83.8 & 90.6 & 89.7 & 91.4 \\
\rowcolor{yellow!15} Visual7W-pointing   & 52.0 & 23.4 & 70.2 & 61.8 & 83.6 & 84.1 & 78.7 & 93.8 \\
\textit{All Visual Grounding} & 47.0 & 26.0 & 58.9 & 60.2 & 82.6 & 85.1 & 82.8  & 90.2 \\
\midrule

\rowcolor{cyan!15} \textbf{Final Score (36 tasks)} & & & & & & & & \\
All                  & 28.0 & 27.1 & 43.7 & 37.5 & 63.3 & 66.6 & 68.0 & 71.2 \\
All IND              & 25.3 & 31.0 & 38.1 & 34.2 & 64.9 & 68.4 & 72.0 & 74.8 \\
All OOD              & 25.1 & 23.7 & 45.6 & 33.4 & 53.9 & 57.9 & 63.0 & 66.7 \\

\bottomrule
\end{tabular}
}
\caption{The comprehensive evaluation results comparing baseline methods with our \mdname\ on the MMEB benchmark, comprising 20 in-distribution and 16 out-of-distribution datasets (the OOD marked with yellow). The \mdname\ uses Qwen2-VL-7B as its backbone, and \mdname$^\dag$ denotes using LLaVA-OneVision-7B as its backbone.}
\label{tab:supp_main_MMEB_exp_per_task}
\end{table*}
Tab.~\ref{tab:supp_main_MMEB_exp_per_task} presents comprehensive results on the MMEB benchmark across eight models: BLIP2, MagicLens, EVA-CLIP, E5-V, VLM2Vec, UniME, \mdname, and \mdname$^\dag$. Results for BLIP2 through UniME are reproduced directly from UniME~\cite{unime}. Our implementation details specify that \mdname\ employs the Qwen2-VL-7B backbone, while \mdname$^\dag$ represents uses LLaVA-OneVision-7B as its backbone.

\section{Further Analysis}
\label{sec:further_analysis}

\subsection{Visualization of the Training Data}

\begin{figure}[t!]
\centering
\includegraphics[width=\linewidth]{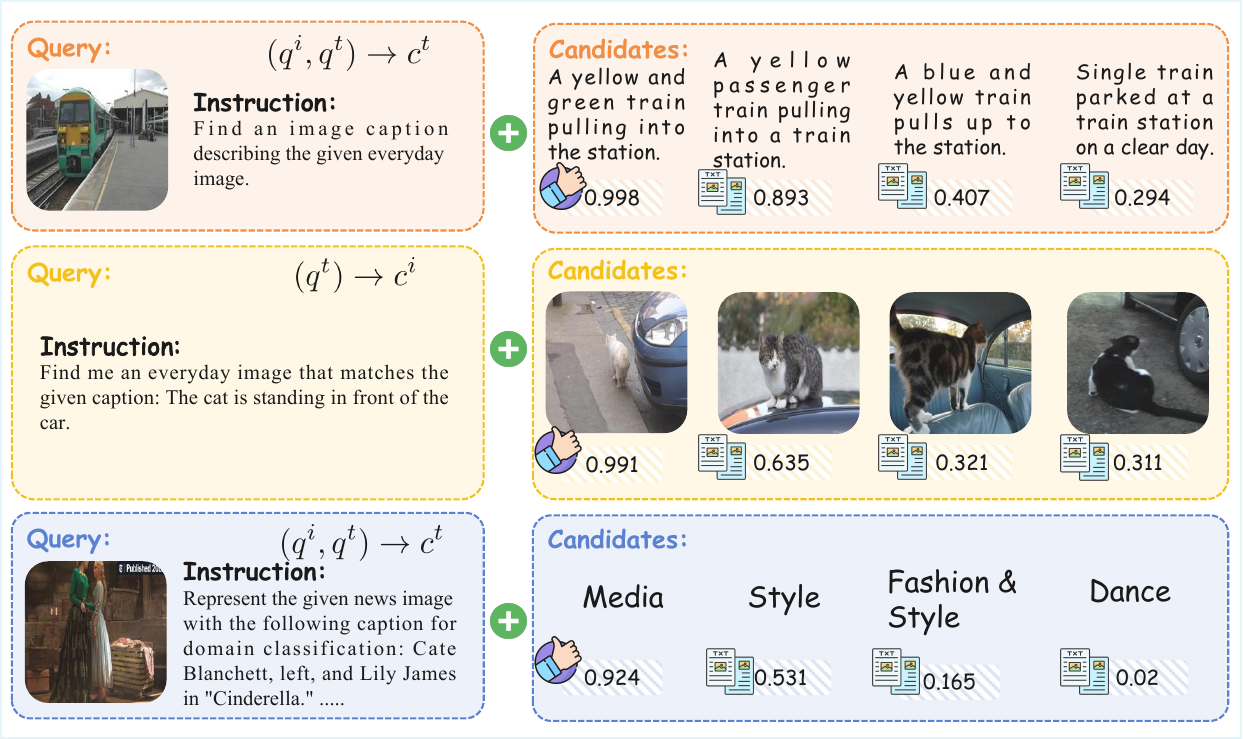}
\caption{Qualitative examples. We present examples showing queries and their corresponding hard negative candidates processed after our hard negative mining pipeline.}
\label{fig:vis1}
\end{figure}

Fig.~\ref{fig:vis1} presents training examples from the MMEB dataset annotated with their semantic matching scores, which obtained after our proposed pipeline. The visualization demonstrates: (1) target candidates achieving the highest match scores (nearly 1.0), (2) partially relevant candidates with intermediate scores (between 0.0 and 1.0), and (3) irrelevant candidates receiving near-zero scores.

\subsection{Visualization of the Retrieval and Rerank Results}

\begin{figure}[t!]
\centering
\includegraphics[width=\linewidth]{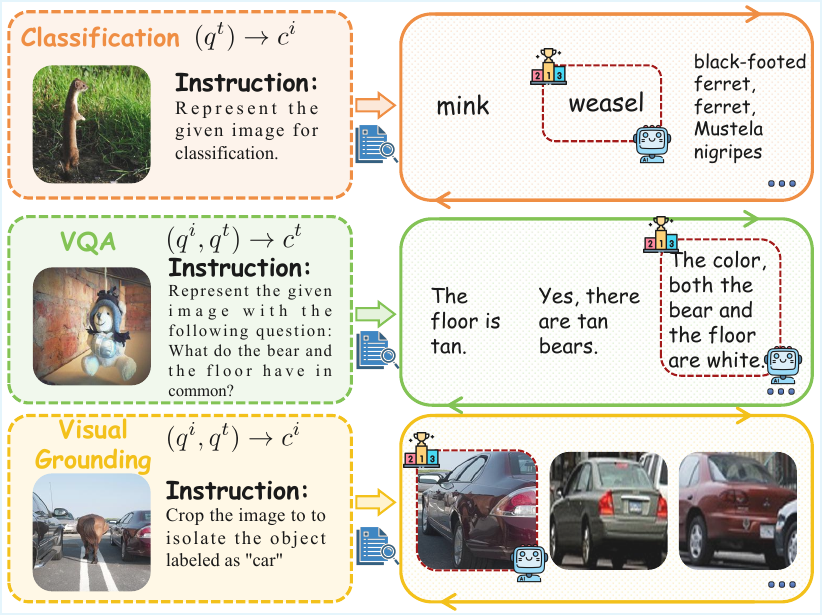}
\caption{Qualitative examples. We present the additional retrieval and reranking results of our method across different tasks.}
\label{fig:vis2}
\end{figure}

Fig.~\ref{fig:vis2} presents additional qualitative results demonstrating our method's performance across multiple tasks. The visualization reveals that while \mdname\ successfully retrieves query-matched candidates, \mdrkname\ further refines these results by selecting the optimally matched candidate as the final output.
\end{document}